\pdfoutput=1
\PassOptionsToPackage{dvipsnames,table}{xcolor}
\documentclass[11pt,letterpaper]{style}

\usepackage[numbers]{natbib}
\usepackage{graphicx}
\usepackage{subcaption}
\usepackage{booktabs}
\usepackage{enumitem}
\usepackage{tabularx}
\usepackage{listings}
\usepackage{amsmath,amsfonts,amssymb}
\usepackage{mathtools}
\usepackage{amsthm}
\usepackage{algorithm}
\usepackage{algorithmic}
\usepackage{multirow}
\usepackage{colortbl}
\usepackage{xurl}
\usepackage{float}
\usepackage{placeins}
\usepackage{tcolorbox}
\tcbuselibrary{skins,breakable,listingsutf8}
\graphicspath{{./}{figures/}{logo/}}

\definecolor{darkblue}{rgb}{0, 0, 0.5}
\definecolor{darkred}{RGB}{180, 30, 30}
\definecolor{midred}{RGB}{210, 80, 80}
\definecolor{lightred}{RGB}{240, 140, 140}
\definecolor{tablehead}{HTML}{F5EFFF}
\definecolor{altrow}{HTML}{FCFAFF}
\definecolor{overallrow}{HTML}{F7F1FF}
\definecolor{missingrow}{HTML}{F0EDF4}
\definecolor{bestgreen}{RGB}{205, 232, 214}
\definecolor{failsoft}{RGB}{252, 234, 232}
\definecolor{failmid}{RGB}{247, 213, 210}
\definecolor{failstrong}{RGB}{239, 190, 187}
\definecolor{modelsep}{HTML}{DDD1F0}

\newcommand{\para}[1]{\textbf{#1}}
\newcommand{\best}[1]{\cellcolor{bestgreen}\textbf{#1}}
\newcommand{\bad}[1]{\cellcolor{failstrong}\textbf{#1}}
\newcommand{\weak}[1]{\cellcolor{failmid}#1}
\newcommand{\lowcell}[1]{\cellcolor{failsoft}#1}

\newcommand{\modelrule}{\arrayrulecolor{modelsep}\specialrule{0.25pt}{0.15ex}{0.15ex}\arrayrulecolor{black}}
\newcommand{\avgse}[2]{#1$_{\pm \text{#2}}$}

\newcommand{\K}{\mathcal{K}}
\newcommand{\agents}{\mathcal{A}}

\newcommand{\mechanics}{\mathcal{M}}
\newcommand{\task}{\mathcal{T}}
\newcommand{\goal}{\varphi}
\newcommand{\taskdesc}{\delta}
\newcommand{\pddl}{\textsc{pddl}}
\newcommand{\passk}[1]{{Pass\textasciicircum#1}}

\newcommand{\shortname}{\texttt{EnactToM}}
\newcommand{\codeus}{\_\allowbreak}

\title{EnactToM: An Evolving Benchmark for Functional Theory of Mind in Embodied Agents}
\runningtitle{EnactToM}

\newcommand{\ucsb}{1}
\newcommand{\kcl}{2}
\newcommand{\cisco}{3}

\author[\ucsb]{Gurusha Juneja\textsuperscript{*}}
\author[\ucsb]{Dylan Lu\textsuperscript{*}}
\author[\ucsb]{Saaket Agashe}
\author[\ucsb]{Parth Diwane}
\author[\kcl]{Edward Gunn}
\author[\cisco]{Jayanth Srinivasa}
\author[\cisco]{Gaowen Liu}
\author[\ucsb]{William Yang Wang}
\author[\kcl]{Yali Du}
\author[\ucsb]{Xin Eric Wang}

\affil[\ucsb]{University of California, Santa Barbara}
\affil[\kcl]{King's College London}
\affil[\cisco]{Cisco Research}

\begin{document}

\begin{abstract}
Theory of Mind (ToM), the ability to track others' epistemic state, makes humans efficient collaborators.
AI agents need the same capacity in multi-agent settings, yet existing benchmarks mostly test \emph{literal} ToM by asking direct belief questions.
The ability act optimally on implicit beliefs in embodied environments, called \emph{functional} ToM, remains largely untested.
We introduce \shortname{}, an evolving benchmark of 300 embodied multi-agent  tasks set in a 3D household with partial observability, private information, and constrained communication.
Each task is formally verified for solvability and required epistemic depth, and new tasks are generated increase difficulty as models improve.
On the hard split, all seven evaluated frontier models score 0.0\% \passk{3} on  functional task completion, while averaging 45.0\% on literal belief probes. Manual analysis traces 93\% of sampled failures to epistemic coordination breakdowns such as withheld information, ignored partner constraints, and misallocated messages, providing a concrete target for future work.
\vspace{3mm}

\parbox{\linewidth}{
\textbf{Correspondence:} \href{mailto:gurusha@ucsb.edu}{\texttt{gurusha@ucsb.edu}}, \href{mailto:ericxwang@ucsb.edu}{\texttt{ericxwang@ucsb.edu}}\\
\textbf{Project Page:} \href{https://enact-tom.github.io/}{\texttt{enact-tom.github.io}}
}
\end{abstract}

\maketitle
\renewcommand{\thefootnote}{*}
\footnotetext{Equal contribution.}
\renewcommand{\thefootnote}{\arabic{footnote}}

\section{Introduction}
\label{sec:introduction}

Language Models (LMs) are increasingly deployed as agents in shared embodied settings where effective collaboration requires Theory of Mind (ToM), the ability to model the beliefs, intentions, and observations of others~\citep{premack1978does, wimmer1983beliefs, tomasello2005understanding}. However, even when frontier LMs can \emph{report} another agent's mental state when prompted (\emph{literal} ToM~\citep{riemer2025positiontheorymindbenchmarks}), they routinely fail to \emph{act} on that knowledge during grounded multi-agent coordination, demonstrating a lack of \emph{functional} ToM.

\begin{figure}[t]
  \centering
  \includegraphics[width=\linewidth]{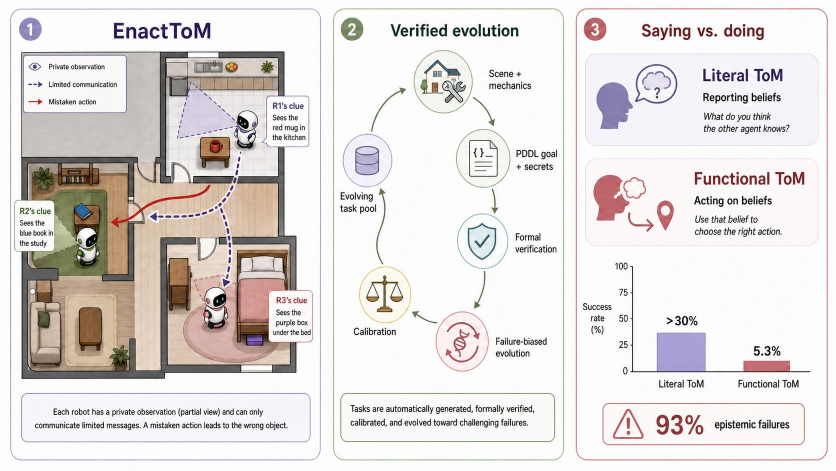}
  \caption{\textbf{EnactToM overview.} \textbf{(1) Embodied task:} agents operate in a shared 3D household environment but receive different private observations and can exchange only limited messages; success requires choosing actions that account for what teammates know and do not know. \textbf{(2) Verified evolution:} tasks are generated from scenes and mechanics, instantiated as PDDL goals with private secrets, checked for formal solvability and epistemic validity, calibrated, and then evolved from model failures to keep the benchmark difficult. \textbf{(3) Saying vs.\ doing:} the evaluation separates literal ToM (reporting another agent's belief when asked) from functional ToM (using that belief to act correctly), revealing that models can often say what others know while still failing to coordinate.}
  \label{fig:main_overview}
\end{figure}

An agent lacking functional ToM is prone to severe coordination failures: it may redundantly broadcast already-known information, fail to anticipate collaborators' actions, or remain idle awaiting unnecessary instructions. In the worst case, this inability to infer intent can lead to critically unsafe actions or actively disrupt the very human environments the system was designed to assist.

Existing benchmarks lack three properties needed to measure this gap. \emph{Functional}: success must depend on actions that use partner beliefs, not by simply answering belief questions. \emph{Grounded}: the evaluation must take place in environments where spatial constraints and information asymmetry arise naturally, reflecting the conditions under which deployed embodied agents will need to exhibit ToM in practice. \emph{Saturation-resistant}: as models improve, the evaluation must generate new tasks against remaining failure modes, with guarantees of solvability and required epistemic depth.

We address these gaps with \shortname{}, an evolving benchmark for functional Theory of Mind in embodied multi-agent settings. Our contributions are: (1)~a \textbf{300-task benchmark} that scores functional task completion and literal belief probes on the same task instances, isolating the act-vs-report gap on a per-task basis; (2)~an \textbf{evolving generation pipeline} in which an autonomous coding agent authors \pddl{}-verified tasks at a target epistemic depth and seeds successive rounds from current frontier-model failures, hardening the pool without modifying the generator; and (3)~an \textbf{analysis of frontier-model behavior} across seven LMs that decomposes 93\% of sampled failures into five recurring epistemic-coordination breakdowns.

On the hard split of \shortname{}, every one of the seven frontier models we evaluate scores 0.0\% \passk{3} for functional task completion. No model coordinates reproducibly across three independent runs, while the same models correctly answer up to 45.0\% of literal belief probes on the same tasks. Figure~\ref{fig:main_overview} summarizes the full evaluation loop: private observations and constrained communication create the need for functional ToM, formal verification keeps generated tasks solvable and epistemically valid, and the final evaluation compares belief reporting against belief-guided action. Models \emph{report} their partners' knowledge but cannot reliably \emph{use} it to act. The five failure modes catalogued in Section~\ref{sec:failure_behaviors} are withheld information, broken epistemic chains, ignored partner constraints, misallocated messages, and sabotaged private incentives. These give concrete targets for follow-up work.

\section{Related Work}

Existing ToM evaluations are text-based and test \emph{literal} ToM only: they prompt agents with explicit questions about others' beliefs and measure whether the answer is correct~\citep{le2019revisiting, wu2023hi, gandhi2024understanding, kim2023fantom, exploretom2024, xu2024opentom, jin-etal-2024-mmtom, mumatom2024, riemer2025positiontheorymindbenchmarks}. FANToM~\citep{kim2023fantom} and OpenToM~\citep{xu2024opentom} scale to multi-party conversations and open-ended generation, but the agent is still an observer answering questions, never a participant who must act on inferred beliefs. Recent work confirms a gap between tracking beliefs and acting on them~\citep{zhou2024far, gu2024simpletom}: models that pass false-belief tests can still fail to use that information when choosing actions. No existing benchmark measures this gap in grounded settings where agents must physically coordinate.

Embodied benchmarks~\citep{szot2021habitat, chang2025partnr, padmakumar2022teach, bard2020hanabi, zhou2024sotopia} test multi-agent coordination but none formally require epistemic reasoning or verify that task success depends on modeling what partners know. PARTNR~\citep{chang2025partnr} evaluates collaborative household tasks in Habitat but assumes shared observability and does not introduce information asymmetry. Hanabi~\citep{bard2020hanabi} requires reasoning about hidden information but is a card game without spatial grounding or communication constraints.

Static benchmarks also saturate as models improve~\citep{akhtar2026saturation, ott2022mapping} and suffer contamination when tasks enter training data~\citep{jacovi2023stop, golchin2024time}. Dynamic approaches like Dynabench~\citep{kiela2021dynabench} and LiveBench~\citep{white2024livebench} address staleness through human-in-the-loop or periodic refreshes, but neither provides formal solvability guarantees or epistemic depth verification. \shortname{} addresses all three gaps: it is embodied, measures functional ToM through action rather than verbal report, and evolves its task pool to stay ahead of improving models. An extended discussion is in Appendix~\ref{app:related_work}; ToM studies from cognitive science are in Appendix~\ref{app:tom-cogsci}.

\section{Preliminaries}
\label{sec:preliminaries}

\subsection{Functional and Literal Theory of Mind}
\label{sec:func_tom}

Literal Theory of Mind evaluations probe an agent's ability to report the beliefs, desires, or intentions of others when asked directly. On the other hand, Functional Theory of Mind requires agents to \emph{act} based on their understanding of others' mental states in order to succeed at a task~\citep{riemer2025positiontheorymindbenchmarks}. In our setting, agents operate in a shared 3D environment with private information, limited communication, and physical constraints. Success in this setting depends not on answering questions about others' beliefs but on choosing actions that account for what others know, can observe, and intend to do.

\subsection{Epistemic operators and ToM depth}
\label{sec:epistemic_ops}

Consider a task in which Agent~A places a bowl on a table and Agent~B must \emph{confirm} the placement. B can only confirm what it has observed or been told, so the goal is not just that the bowl is on the table but that B \emph{knows} it is. If a third agent C must verify that B knows then it must reason about B's knowledge. Each additional layer demands a deeper level of Theory of Mind.

We formalize this with the epistemic operator $\K$. $\K_i(\phi)$ asserts that agent $a_i$ knows fact $\phi$. The assertion is satisfied when $a_i$ has directly observed $\phi$ or received it through communication. We say that a task requires depth $d-1$ ToM reasoning when the task success depends on the truth value of the operators nested to depth $d$:
\begin{equation}
  \underbrace{\K_{a_1}(\K_{a_2}(\cdots \K_{a_d}(\phi) \cdots))}_{\text{depth } d}
  \label{eq:tom_depth}
\end{equation}
 At depth~1, an agent needs to learn a fact (zeroth-order). At depth~2, it must reason about what another agent knows (first-order). This connects to level-$k$ reasoning from behavioral game theory~\citep{stahl1994experimental, camerer2004cognitive}, where a level-$k$ agent selects actions assuming others reason at level $k{-}1$. Table~\ref{tab:tom_orders} summarizes the orders used throughout the benchmark.

\begin{table}[t]
\centering
\caption{Orders of Theory of Mind used to author and validate \shortname{} tasks. The reported benchmark caps generated tasks at depth~3.}
\label{tab:tom_orders}
\small
\renewcommand{\arraystretch}{1.08}
\setlength{\tabcolsep}{2.5pt}
\begin{tabularx}{\linewidth}{@{} l l X @{}}
\toprule
\textbf{Order} & \textbf{Pattern} & \textbf{\shortname{} meaning} \\
\midrule
0: no ToM & $\phi$ & Direct physical goal; no partner knowledge matters. \\
1: self-aware & $\K_a(\phi)$ & Notice one's own information gap and obtain or communicate the missing fact. \\
2: other-aware & $\K_a(\K_b(\phi))$ & Act from a model of what a partner knows and still needs to know. \\
3: recursive & $\K_a(\K_b(\K_c(\phi)))$ & Sustain an epistemic relay over who knows that another agent knows. \\
4+: self-ref. & $\K_a(\K_b(\cdots))$ & Deeper belief loops; excluded because coordination becomes brittle even for humans. \\
\bottomrule
\end{tabularx}
\end{table}

\subsection{Task goals in \pddl{}}
\label{sec:epistemic_pddl}

Each task goal is specified in \pddl{}, combining physical predicates (\texttt{is\_on\_top}, \texttt{is\_open}) with epistemic $\K$-operators. The $\K$-depth of the goal is computed from the nesting structure. The physical predicates determine task success; the $\K$-operators define literal ToM probes that are evaluated separately at the end of each episode (Section~\ref{sec:experiments}). An example goal and the full compilation procedure are in Appendix~\ref{app:compilation}.

\subsection{Mechanics}
\label{sec:mechanics}

Mechanics model constraints that arise in real coordination. \textbf{Room restriction} confines agents to a subset of rooms, like teams operating in different physical spaces. \textbf{Limited bandwidth} caps each agent's messages, modeling channels with limited capacity. \textbf{Restricted communication} permits messages only along a fixed graph, modeling networks where not every pair of agents can communicate directly. \textbf{Remote control}, \textbf{state mirroring}, and \textbf{inverse state} all model effects the actor cannot observe directly: a trigger object actuating a target across rooms, two objects' states staying synchronized, and an affordance being reversed. The generation agent draws and composes any subset; Appendix~\ref{app:mechanics_grounding} expands the groundings.

\section{The EnactToM Framework}
\label{sec:methodology}

Evaluating functional Theory of Mind requires tasks where success depends on agents adapting their actions to the private knowledge, access, and intentions of other agents, not merely reporting beliefs when prompted. Hand-crafting such tasks does not scale, and as models improve, fixed benchmarks saturate. We propose an agentic task generation framework that addresses both problems. An autonomous coding agent authors multi-agent ToM tasks inside a sandboxed workspace, invoking verifiers that ensure each task is logically solvable, physically executable, and genuinely requires epistemic reasoning.

\subsection{Task representation}
\label{sec:task_repr}

Each task $\task$ is a tuple:
\begin{equation}
  \task = \bigl(\mathcal{S},\; \agents,\; \goal,\; \taskdesc,\; \mechanics,\; \Sigma,\; \mathcal{C} \bigr)
  \label{eq:task_tuple}
\end{equation}
where $\mathcal{S}$ is an HSSD scene~\citep{szot2021habitat}, $\agents = \{a_1, \ldots, a_n\}$ is a set of agents with spawn positions, $\goal$ is a \pddl{} goal formula combining physical predicates with $\K$-operators (Section~\ref{sec:epistemic_ops}), $\taskdesc$ is a natural-language task description shared with all agents, $\mechanics$ is a set of mechanic bindings drawn from the registry (Section~\ref{sec:mechanics}), $\Sigma$ maps each agent to private secrets, and $\mathcal{C} \in \{\text{cooperative}, \text{mixed}\}$ is the task category.

\para{Secrets.} Each agent receives private secrets $\Sigma(a_i)$: natural-language statements of room restrictions, target object identities, and mechanic hints. Secrets state \emph{what} each agent privately knows but never \emph{how} to coordinate. Figuring out who to tell, what to say, and when to act is the epistemic coordination challenge the benchmark measures.

\para{Categories.} Each category targets a different facet of epistemic reasoning. A task is considered solved when all physical predicates in $\goal$ are satisfied by the end of the episode.

\emph{Cooperative} tasks give all agents a shared $\goal$. Agents have different room access and private knowledge. Success requires recognizing what partners cannot observe and communicating accordingly. This tests \textbf{epistemic perspective-taking}: can an agent infer what its partner does not know and act on that inference?

\emph{Mixed-motive} tasks combine a shared cooperative goal with private per-agent side-objectives. The private objectives are not in direct conflict with the shared goal, but they introduce additional coordination demands: agents must allocate limited turns and messages across shared and private tasks, and must reason about whether partners are spending effort on private objectives that could delay the shared plan. This tests \textbf{strategic resource allocation under partial information about partner priorities}.

\begin{algorithm}[t]
\caption{\shortname{} task generation.}
\label{alg:taskgen}
{\small
\begin{algorithmic}[1]
\STATE \textbf{Input:} category $\mathcal{C}$, target depth $d$, seed pool $\mathcal{P}$, seed-task failure ratio $\rho$
\STATE \textbf{Output:} accepted task $\task$
\STATE $\mathcal{S} \gets \texttt{new\_scene}()$
\STATE $\mathcal{P}_{\text{seed}} \gets \textsc{SampleSeedTasks}(\mathcal{P}, \rho)$
\STATE $\task \gets \textsc{AuthorTask}(\mathcal{S}, \mathcal{P}_{\text{seed}}, \mathcal{C}, d)$
\WHILE{\texttt{judge}($\task$) or \texttt{test\_task}($\task$) rejects}
  \STATE $\task \gets \textsc{ReviseTask}(\task)$
\ENDWHILE
\STATE \textbf{return} \texttt{submit\_task}($\task$)
\end{algorithmic}
}
\end{algorithm}

\subsection{Task generation agent}
\label{sec:gen_agent}

We use the \texttt{mini-SWE-agent} \citep{yang2024sweagent} harness in a sandboxed workspace seeded with the scene graph, a blank task template, reference files listing available mechanics, predicates, and actions, and in-context seed tasks drawn from the existing pool with failure ratio $\rho$ (Section~\ref{sec:calibration}; Appendix~\ref{app:workspace}). The agent edits the task file with \texttt{bash} and \texttt{jq} and invokes five tools: \texttt{new\_scene}, \texttt{bash}, \texttt{judge}, \texttt{test\_task}, and \texttt{submit\_task}. The only human inputs are the target category $\mathcal{C} \in \{\text{cooperative}, \text{mixed}\}$ and ToM depth $d \in \{1,2,3\}$; authoring, evaluation, and revision are otherwise end-to-end.

\para{Authoring order.} The agent writes the formal \pddl{} goal $\goal$ first, then derives $\taskdesc$ and per-agent secrets $\Sigma(a_i)$ from it. Anchoring the narrative to $\goal$ prevents drift between the description and the formal specification. Secrets carry only constraints, target IDs, and mechanic hints; the prompt forbids encoding coordination strategy, which the agents must work out themselves.

\para{Iteration.} The agent invokes \texttt{judge} and \texttt{test\_task} in any order and may revise goals, swap mechanics, or call \texttt{new\_scene} to restart. Tools return structured feedback: per-criterion scores with required fixes, planner diagnostics, or pass/fail traces. The agent uses this feedback to revise. \texttt{submit\_task} requires both \texttt{judge} and \texttt{test\_task} to have passed.

\subsection{Verifiers}

\para{PDDL Parsing.}
\label{sec:verification}
The \texttt{judge} tool first runs two deterministic checks: structural validation that the \pddl{} goal is syntactically valid, all referenced objects and furniture exist in the scene, and mechanic bindings are well-formed; and a $\K$-depth check that rejects the task when the nesting depth of epistemic operators in $\goal$ (Section~\ref{sec:epistemic_pddl}) does not match the requested target $d$.\footnote{See Appendix~\ref{app:design_decisions} for why the planner is not given to the evaluated agent as a tool.} These cheap checks gate the more expensive judge and calibration runs; physical executability is verified separately by the calibrator below.

\para{LLM Judge Council.}
\label{sec:judge}
Two LLMs (Kimi-K2.5 and GPT-5.2) independently score each candidate on eight $[0,1]$ criteria, and a task passes only when both agree. Seven criteria are shared: \textbf{agent necessity}, \textbf{secret quality}, \textbf{public/private grounding}, \textbf{narrative consistency}, \textbf{goal relevance}, \textbf{mechanic utilization}, and \textbf{formal goal quality}; the eighth is category-specific: \textbf{task interdependence} (cooperative) or \textbf{subgoal tension} (mixed). Acceptance requires mean $\bar{s}\geq 0.65$ with a per-criterion floor $s_c\geq 0.5$, both tuned on 50 manually rated tasks. On failure, the council returns per-criterion scores and concrete fixes (e.g., ``add a secret for agent\_1 explaining the limited\_bandwidth mechanic'') that drive the next revision. Appendices~\ref{app:judge_prompt} and~\ref{app:design_decisions} give the full prompt, sample feedback, and rationale for each criterion.

\para{Structural calibrator.}
\label{sec:tom_necessity}
Success on \shortname{} depends on two factors: embodied reasoning (navigation, object recognition, manipulation) and Theory of Mind. To isolate ToM, the calibrator runs each candidate in a \emph{baseline} condition with all secrets revealed to every agent. If agents fail with full knowledge, the gap is embodied and the task is rejected.\footnote{This is just a baseline for eliminating environmental and embodiment failures. We do not see the solution of this problem to look like this.} Only tasks that pass the baseline enter the benchmark, so scores reflect ToM.

\subsection{Difficulty evolution}
\label{sec:calibration}

A fixed benchmark saturates as models improve. \shortname{} addresses this by evolving the task pool. The generation agent receives seed tasks from the existing pool as in-context examples (Section~\ref{sec:gen_agent}), sampled by their performance against a family of target models (GPT-5.4, Sonnet, DeepSeek-v3.2): a fraction $\rho$ are tasks these models failed, and the remainder are tasks they passed. We refer to $\rho$ as the \emph{seed-task failure ratio}; higher $\rho$ pushes the generator toward the epistemic coordination patterns current frontier models fail on. As new tasks enter the pool and are benchmarked, the seed distribution shifts. Later generation runs see harder examples, creating evolutionary pressure without changing the generation infrastructure.

\section{Experiments}
\label{sec:experiments}

\para{Experimental setup.} We evaluate seven frontier models on two \shortname{} splits, \emph{standard} and \emph{hard}, each containing 150 tasks spanning cooperative and mixed-motive categories. The splits differ only in seed-task failure ratio during generation, $\rho = 0.8$ for standard, $\rho = 0.9$ for hard. The hard split concentrates on tasks current frontier models fail. Full pool composition is in Table~\ref{tab:dataset_stats}.\footnote{We limit to depth~3 because reasoning beyond $d{=}3$ is difficult even for humans.} Each episode is given double the calibration-baseline turn and message budget to account for coordination overhead of partial information.

\para{Success criteria.} A task is solved once all physical predicates in $\goal$ are completed; this is the \emph{functional} measure. Separately, the $\K$-operators in $\goal$ define \emph{literal ToM probes}: at episode end each agent is asked an explicit question about another agent's knowledge state (e.g., ``what state does agent\_0 think the fridge is in?''); the full probe prompt is in Appendix~\ref{app:literal_tom_prompt}. The probe accuracy, reported as ``Literal'' in Table~\ref{tab:main_results}, measures whether agents can \emph{report} beliefs, independent of whether they \emph{acted} on them.

\para{Metrics.} Each task is run $n{=}3$ times. \textbf{Avg} is the mean per-run pass rate, reported with binomial standard error over fixed attempts. \textbf{Pass@3} is a union metric: success if at least one of the three runs succeeds. \textbf{\passk{3}} is an exact metric: success only if all three runs succeed. Avg shows single-attempt capability and Pass@3 the best-of-3 ceiling, but a one-in-three success is luck, not ToM. We emphasize \passk{3}: coordination should be reproducible.

\begin{table}[H]
\centering
\caption{\shortname{} results on matched standard and hard subsets. Each model is evaluated over cooperative, mixed, and overall scopes; each split reports functional task success and literal ToM probe success. Definitions of Avg, Pass@3, and \passk{3} are given in the experimental setup. Rose cells emphasize low performance, darker rose marks exact zero, and green marks the best overall exact \passk{3} in each split and metric family. Asterisks mark partial API runs; missing attempts are counted as non-passes under fixed $n{=}3$ accounting.}
\label{tab:main_results}
\begingroup
\fontsize{5.0pt}{5.25pt}\selectfont
\renewcommand{\arraystretch}{1.0}
\setlength{\tabcolsep}{3.6pt}
\rowcolors{5}{white}{altrow}
\resizebox{\linewidth}{!}{%
\begin{tabular}{@{} ll *{12}{r} @{}}
\toprule
% \rowcolor{tablehead}
\textbf{Model} & \textbf{Scope}
& \multicolumn{6}{c}{\textbf{Standard}}
& \multicolumn{6}{c}{\textbf{Hard}} \\
\cmidrule(lr){3-8} \cmidrule(l){9-14}
% \rowcolor{tablehead}
&& \multicolumn{3}{c}{\textbf{Functional}} & \multicolumn{3}{c}{\textbf{Literal}}
& \multicolumn{3}{c}{\textbf{Functional}} & \multicolumn{3}{c}{\textbf{Literal}} \\
\cmidrule(lr){3-5} \cmidrule(lr){6-8} \cmidrule(lr){9-11} \cmidrule(l){12-14}
% \rowcolor{tablehead}
&& \textbf{Avg} & \textbf{Pass@3} & \textbf{\passk{3}}
& \textbf{Avg} & \textbf{Pass@3} & \textbf{\passk{3}}
& \textbf{Avg} & \textbf{Pass@3} & \textbf{\passk{3}}
& \textbf{Avg} & \textbf{Pass@3} & \textbf{\passk{3}} \\
\midrule
Gemini-Pro & Coop & \avgse{53.0}{6.1} & 77.3 & 22.7 & \avgse{34.8}{5.9} & 59.1 & \lowcell{13.6} & \lowcell{\avgse{7.6}{3.3}} & 22.7 & \bad{0.0} & \avgse{60.6}{6.0} & 86.4 & 36.4 \\
& Mixed & \avgse{22.2}{5.7} & 44.4 & \bad{0.0} & \avgse{18.5}{5.3} & 33.3 & \lowcell{11.1} & \weak{\avgse{5.6}{3.1}} & 16.7 & \bad{0.0} & \avgse{66.7}{6.4} & 88.9 & 38.9 \\
\rowcolor{overallrow} & \textbf{Overall} & \avgse{39.2}{4.5} & 62.5 & \lowcell{12.5} & \avgse{27.5}{4.1} & 47.5 & \lowcell{12.5} & \lowcell{\avgse{6.7}{2.3}} & 20.0 & \bad{0.0} & \avgse{63.3}{4.4} & 87.5 & \best{37.5} \\
\modelrule
Gemini-Flash & Coop & \avgse{43.9}{6.1} & 72.7 & 22.7 & \avgse{39.4}{6.0} & 72.7 & 18.2 & \weak{\avgse{3.0}{2.1}} & \lowcell{9.1} & \bad{0.0} & \avgse{45.5}{6.1} & 72.7 & \lowcell{13.6} \\
& Mixed & \avgse{40.7}{6.7} & 66.7 & 22.2 & \lowcell{\avgse{13.0}{4.6}} & 27.8 & \weak{5.6} & \weak{\avgse{5.6}{3.1}} & 16.7 & \bad{0.0} & \avgse{38.9}{6.6} & 72.2 & \lowcell{11.1} \\
\rowcolor{overallrow} & \textbf{Overall} & \avgse{42.5}{4.5} & 70.0 & \best{22.5} & \avgse{27.5}{4.1} & 52.5 & \lowcell{12.5} & \weak{\avgse{4.2}{1.8}} & \lowcell{12.5} & \bad{0.0} & \avgse{42.5}{4.5} & 72.5 & \lowcell{12.5} \\
\modelrule
GPT-5.4 & Coop & \avgse{21.2}{5.0} & 36.4 & \lowcell{9.1} & \avgse{22.7}{5.2} & 45.5 & \bad{0.0} & \weak{\avgse{3.0}{2.1}} & \lowcell{9.1} & \bad{0.0} & \avgse{45.5}{6.1} & 72.7 & \lowcell{13.6} \\
& Mixed & \lowcell{\avgse{13.0}{4.6}} & 27.8 & \bad{0.0} & \lowcell{\avgse{11.1}{4.3}} & 22.2 & \bad{0.0} & \weak{\avgse{3.7}{2.6}} & \lowcell{11.1} & \bad{0.0} & \avgse{42.6}{6.7} & 83.3 & 16.7 \\
\rowcolor{overallrow} & \textbf{Overall} & \avgse{17.5}{3.5} & 32.5 & \weak{5.0} & \avgse{17.5}{3.5} & 35.0 & \bad{0.0} & \weak{\avgse{3.3}{1.6}} & \lowcell{10.0} & \bad{0.0} & \avgse{44.2}{4.5} & 77.5 & 15.0 \\
\modelrule
O3 & Coop & \lowcell{\avgse{13.6}{4.2}} & 31.8 & \bad{0.0} & \avgse{33.3}{5.8} & 59.1 & \lowcell{13.6} & \lowcell{\avgse{7.6}{3.3}} & 22.7 & \bad{0.0} & \avgse{56.1}{6.1} & 86.4 & 31.8 \\
& Mixed & \lowcell{\avgse{11.1}{4.3}} & 22.2 & \bad{0.0} & \avgse{35.2}{6.5} & 50.0 & 16.7 & \weak{\avgse{1.9}{1.8}} & \weak{5.6} & \bad{0.0} & \avgse{48.1}{6.8} & 88.9 & 16.7 \\
\rowcolor{overallrow} & \textbf{Overall} & \lowcell{\avgse{12.5}{3.0}} & 27.5 & \bad{0.0} & \avgse{34.2}{4.3} & 55.0 & \best{15.0} & \weak{\avgse{5.0}{2.0}} & 15.0 & \bad{0.0} & \avgse{52.5}{4.6} & 87.5 & 25.0 \\
\modelrule
Kimi-K2.5$^\ast$ & Coop & \lowcell{\avgse{6.1}{2.9}} & \lowcell{13.6} & \bad{0.0} & \lowcell{\avgse{6.1}{2.9}} & \lowcell{9.1} & \bad{0.0} & \weak{\avgse{3.0}{2.1}} & \weak{4.5} & \bad{0.0} & \avgse{42.4}{6.1} & 68.2 & 18.2 \\
& Mixed & \weak{\avgse{5.6}{3.1}} & 16.7 & \bad{0.0} & \weak{\avgse{3.7}{2.6}} & \lowcell{11.1} & \bad{0.0} & \lowcell{\avgse{9.3}{3.9}} & 27.8 & \bad{0.0} & \avgse{46.3}{6.8} & 72.2 & 16.7 \\
\rowcolor{overallrow} & \textbf{Overall} & \weak{\avgse{5.8}{2.1}} & 15.0 & \bad{0.0} & \weak{\avgse{5.0}{2.0}} & \lowcell{10.0} & \bad{0.0} & \weak{\avgse{5.8}{2.1}} & 15.0 & \bad{0.0} & \avgse{44.2}{4.5} & 70.0 & 17.5 \\
\modelrule
GPT-5.4-mini$^\ast$ & Coop & \lowcell{\avgse{10.6}{3.8}} & 18.2 & \bad{0.0} & \avgse{24.2}{5.3} & 54.5 & \weak{4.5} & \weak{\avgse{4.5}{2.6}} & \lowcell{13.6} & \bad{0.0} & \avgse{36.4}{5.9} & 63.6 & \weak{4.5} \\
& Mixed & \lowcell{\avgse{11.1}{4.3}} & 27.8 & \bad{0.0} & \avgse{18.5}{5.3} & 38.9 & \bad{0.0} & \weak{\avgse{1.9}{1.8}} & \weak{5.6} & \bad{0.0} & \avgse{25.9}{6.0} & 61.1 & \bad{0.0} \\
\rowcolor{overallrow} & \textbf{Overall} & \lowcell{\avgse{10.8}{2.8}} & 22.5 & \bad{0.0} & \avgse{21.7}{3.8} & 47.5 & \weak{2.5} & \weak{\avgse{3.3}{1.6}} & \lowcell{10.0} & \bad{0.0} & \avgse{31.7}{4.2} & 62.5 & \weak{2.5} \\
\modelrule
DeepSeek-v3.2$^\ast$ & Coop & \weak{\avgse{3.0}{2.1}} & \lowcell{9.1} & \bad{0.0} & \bad{\avgse{0.0}{0.0}} & \bad{0.0} & \bad{0.0} & \lowcell{\avgse{7.6}{3.3}} & 18.2 & \bad{0.0} & \avgse{34.8}{5.9} & 63.6 & \weak{4.5} \\
& Mixed & \weak{\avgse{1.9}{1.8}} & \weak{5.6} & \bad{0.0} & \bad{\avgse{0.0}{0.0}} & \bad{0.0} & \bad{0.0} & \lowcell{\avgse{9.3}{3.9}} & 27.8 & \bad{0.0} & \avgse{38.9}{6.6} & 66.7 & \weak{5.6} \\
\rowcolor{overallrow} & \textbf{Overall} & \weak{\avgse{2.5}{1.4}} & \lowcell{7.5} & \bad{0.0} & \bad{\avgse{0.0}{0.0}} & \bad{0.0} & \bad{0.0} & \lowcell{\avgse{8.3}{2.5}} & 22.5 & \bad{0.0} & \avgse{36.7}{4.4} & 65.0 & \weak{5.0} \\
\bottomrule
\end{tabular}%
}
\endgroup
\vspace{-1em}
\end{table}

\begin{table}[t]
\centering
\caption{\shortname{} dataset statistics across the 300-task pool.}
\label{tab:dataset_stats}
\small
\setlength{\tabcolsep}{6pt}
\renewcommand{\arraystretch}{1.08}
\begin{tabular}{@{} l r @{}}
\toprule
\textbf{Composition} & \textbf{Count} \\
\midrule
Total tasks            & 300       \\
Cooperative / Mixed    & 150 / 150 \\
2 agents               & 78 (26.0\%) \\
3 agents               & 220 (73.3\%)\\
4+ agents              & 2 (0.7\%)   \\
$\K$-depth 1           & 112 (37.3\%)\\
$\K$-depth 2           & 79 (26.3\%) \\
$\K$-depth 3           & 109 (36.3\%)\\
\midrule
\textbf{Mechanic coverage} & \textbf{Count} \\
\midrule
Room restriction         & 300 (100\%) \\
Limited bandwidth        & 300 (100\%) \\
Restricted communication & 172 (57.3\%)  \\
Remote control           & 48 (16.0\%)   \\
Inverse state            & 9 (3.0\%)     \\
State mirroring          & 5 (1.7\%)     \\
Baseline turns (mean)    & 9.6         \\
Standard turns (mean)    & 19.2 (2$\times$) \\
\bottomrule
\end{tabular}
\end{table}
\section{Results}

\begin{figure}[t]
\centering
\includegraphics[width=\linewidth]{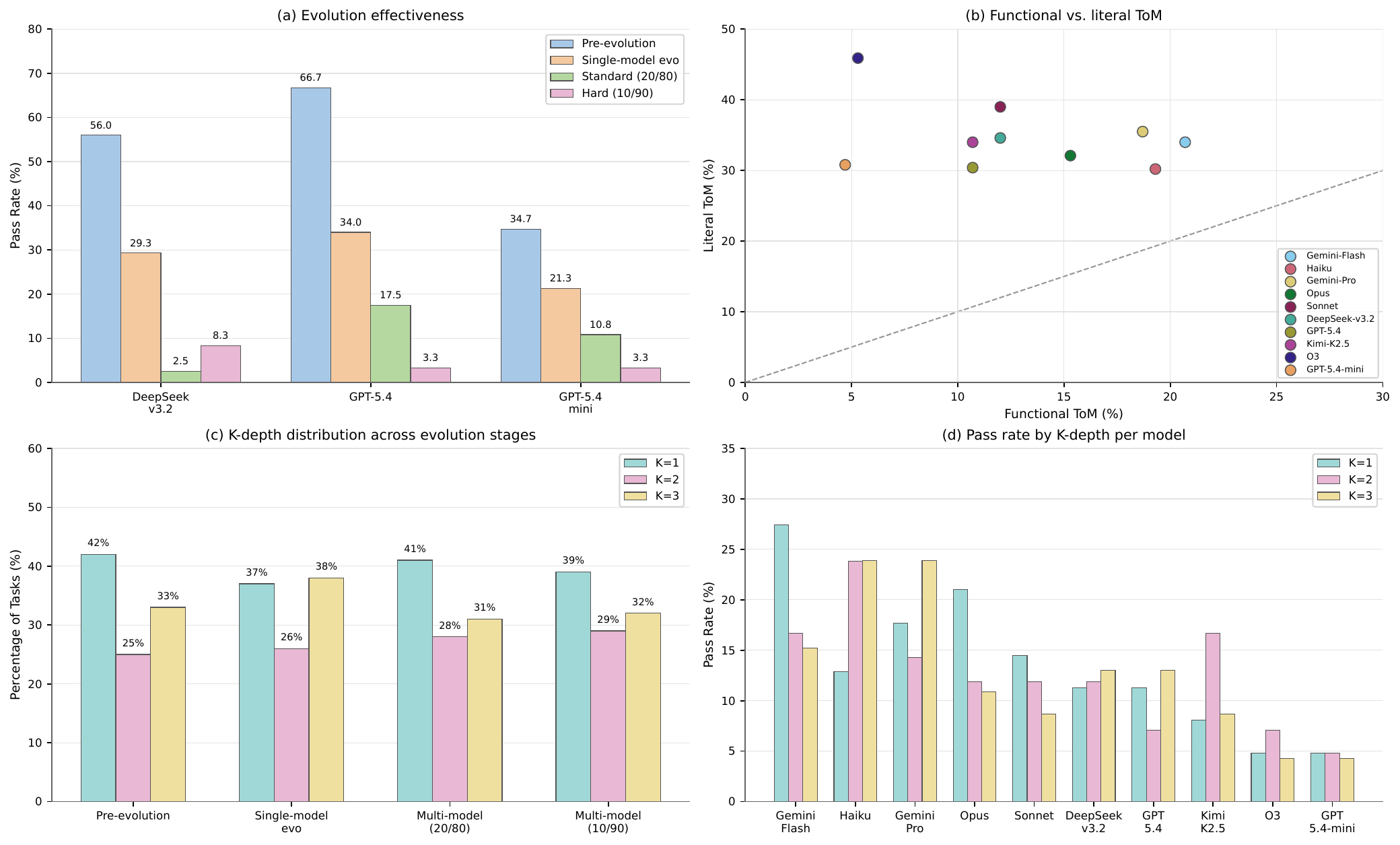}
\caption{\textbf{(a)} Functional Avg single-run pass rate for the three seed models across
pre-evolution, single-model evolution, and multi-model 20/80 and 10/90 pools. \textbf{(b)}
Functional Avg vs.\ literal Avg, showing belief probes exceed embodied task success. \textbf{(c)}
Task percentage at each $\K$ depth across evolution stages, showing hardness is not just deeper
nesting. \textbf{(d)} Functional Avg pass rate by $\K$ depth for each model, showing brittleness at
every depth. Panels (a), (b), and (d) report Avg, not Pass@k or \passk{k}.}
\label{fig:combined_analysis}
\end{figure}

Table~\ref{tab:main_results} gives the main comparison between functional task success and literal belief-probe accuracy on matched \shortname{}-Standard and \shortname{}-Hard tasks. The central pattern is a sharp act--report gap. On the hard split, all seven frontier models achieve 0.0\% overall functional \passk{3}, while their literal Avg scores average 45.0\%. Even the strongest literal models, Gemini-Pro and O3, answer many belief probes correctly but do not convert those beliefs into reproducible coordination. Figure~\ref{fig:combined_analysis} summarizes the evolution, literal--functional gap, and depth analyses. We organize the results around four questions.

\subsection{Does high literal ToM imply high functional ToM?} We find that it does not. On the hard split, every evaluated model scores 0.0\% overall functional \passk{3}, while literal \passk{3} reaches 37.5\% for Gemini-Pro and 25.0\% for O3. The same pattern appears in single-run averages: Gemini-Pro reaches \avgse{63.3}{4.4}\% literal Avg but only \avgse{6.7}{2.3}\% functional Avg on hard tasks, and O3 reaches \avgse{52.5}{4.6}\% literal Avg but only \avgse{5.0}{2.0}\% functional Avg. This gap shows significant separation between reporting a belief and acting on it.

Thus literal ToM probes can overestimate the ToM abilities of these models, suggesting that existing benchmarks~\citep{mumatom2024, jin-etal-2024-mmtom, wu2023hi} do not fully predict behavior in action-constrained settings. In \shortname{}, the useful unit of ToM is the policy that decides when to communicate, whom to inform, and how to route action through constrained partners.

\subsection{Does evolution make tasks more difficult?} We find that evolution does successfully create tasks where reproducible functional success is harder to achieve. The standard subset still admits some stable coordination: Gemini-Flash reaches 22.5\% overall functional \passk{3}, Gemini-Pro reaches 12.5\%, and GPT-5.4 reaches 5.0\%. On the hard subset, every model drops to 0.0\% overall functional \passk{3}. Average functional pass rates also fall from standard to hard for Gemini-Pro (\avgse{39.2}{4.5}\% to \avgse{6.7}{2.3}\%), Gemini-Flash (\avgse{42.5}{4.5}\% to \avgse{4.2}{1.8}\%), GPT-5.4 (\avgse{17.5}{3.5}\% to \avgse{3.3}{1.6}\%), O3 (\avgse{12.5}{3.0}\% to \avgse{5.0}{2.0}\%), and GPT-5.4-mini (\avgse{10.8}{2.8}\% to \avgse{3.3}{1.6}\%). Kimi-K2.5 and DeepSeek-v3.2 also have zero hard functional \passk{3}.

% This shows that The hard split is not merely changing surface form: it uses the same household setting, action space, and verification pipeline, but seeds generation with coordination patterns current models fail. The resulting difficulty transfers across model families.

\subsection{What specific model behavior explains the gap?}
\label{sec:failure_behaviors}
In qualitative analysis, we find that models fail at the operational steps that make beliefs useful. The strongest standard functional model is Gemini-Flash (42.5\% Avg, 22.5\% \passk{3}); the strongest hard literal model is Gemini-Pro (63.3\% Avg, 37.5\% \passk{3}). This shows that models that can answer belief probes still fail to prioritize information relay, reason over partner constraints, or preserve message budget. A manual audit of 40 sampled failures found that 37 were epistemic coordination breakdowns.

Table~\ref{tab:failure_summary} shows that the failures are ToM failures not random simulator mistakes. Agents possess decisive facts but communicate them too late, complete actions without ensuring partners know, or spend scarce messages on low-priority recipients. Current models reason about these ingredients locally, but they do not maintain them as a global coordination state.

\begin{table}[t]
\centering
\footnotesize
\renewcommand{\arraystretch}{0.98}
\begin{tabularx}{\linewidth}{@{} >{\raggedright\arraybackslash}p{0.23\linewidth} >{\raggedright\arraybackslash}p{0.35\linewidth} >{\raggedright\arraybackslash}X @{}}
\toprule
\textbf{Failure pattern} & \textbf{Behavior in the audit} & \textbf{Representative evidence} \\
\midrule
Withholding critical information & In 7 of 40 cases, an agent holds a target, room, or object fact that a partner needs but communicates it only after the partner has already acted on a wrong guess. & \emph{Inspection Staging with a Hidden Target Cabinet}: the target cabinet ID is sent after partners have spent their messages and placed the object on the wrong cabinet (Appendix~\ref{app:fail_relay}). \\
Epistemic chain breakdown & In 8 of 40 cases, an agent completes the physical action but never establishes that the teammate who needs the fact knows it. & \emph{Staging + Fridge Verification}: the fridge is opened, but the opening is never relayed to the agents whose success depends on knowing it (Appendix~\ref{app:fail_kchain}). \\
Private objective sabotage or disclosure & In mixed-motive episodes, agents either damage the shared plan for private gain or reveal private objectives so early that partners can block them. & \emph{Safety Staging with Conflicting Incentives}: one agent announces and executes a fridge-state conflict rather than modeling how teammates will respond (Appendix~\ref{app:fail_mixed}). \\
Misallocating scarce messages & In 4 of 40 cases, agents spend limited messages on the wrong recipient, an unreachable recipient, or low-priority content. & \emph{One-Shot Relay}: the only agent with the critical table ID first messages a blocked recipient, then sends an irrelevant fact to the reachable partner (Appendix~\ref{app:fail_topology}). \\
Ignoring partner constraints & Agents delegate actions to partners who are barred from the relevant room or already constrained by object possession. & \emph{Inspection Prep with Nested Confirmation}: a partner is repeatedly asked to act in a room they cannot enter, wasting two messages before the delegation is corrected (Appendix~\ref{app:fail_constraints}). \\
\bottomrule
\end{tabularx}
\caption{Failure-analysis summary. Full trajectories and quoted actions are in Appendix~\ref{app:tom_failures}.}
\label{tab:failure_summary}
\end{table}

\subsection{How do models behave in cooperative vs.\ strategic settings?} Interestingly, we find that mixed-motive tasks do not uniformly reduce performance relative to cooperative tasks. On hard functional Avg, Kimi-K2.5 is higher on mixed than cooperative tasks (9.3\% vs.\ 3.0\%), as are Gemini-Flash (5.6\% vs.\ 3.0\%), GPT-5.4 (3.7\% vs.\ 3.0\%), and DeepSeek-v3.2 (9.3\% vs.\ 7.6\%). Gemini-Pro, O3, and GPT-5.4-mini show the opposite pattern. Strategic private objectives can either structure action or create new opportunities for premature disclosure and sabotage.

The traces suggest that mixed-motive tasks fail in a different way from cooperative tasks. In cooperative failures, agents usually lose because a useful fact never reaches the right partner. In mixed-motive failures, the model also has to decide whether a fact should be shared at all. Some agents reveal their private goal immediately, making it easy for partners to block or undo it; others overcommit to the private goal and damage the shared objective before the team has established the necessary state. This makes the higher mixed scores for some models informative: private incentives can give the model a clearer local plan, but success requires controlling disclosure while still preserving enough trust and coordination to finish the common task.

\para{K-depth validity.} We manually verify 50 randomly sampled tasks; 49 (98\%) have valid $\K$-depth (Appendix~\ref{app:kdepth_validity}).

\para{Ablations.}
\label{sec:ablations}
We ablate the generation pipeline (Table~\ref{tab:ablations}, Appendix~\ref{app:ablations}). Without \textbf{baseline calibration}, 51\% of tasks are physically unsolvable even with full information. Without \textbf{ICL seed examples}, acceptance drops to 50\% and depth decreases. \textbf{Relaxing the secret quality check} inflates pass rate from 26.7\% to 43.5\% because agents follow prescribed coordination rather than modeling partner knowledge.

\section{Discussion}
\label{sec:discussion}

\para{Limitations and future directions.}
\label{sec:limitations}
\shortname{} caps ToM depth at $d=3$, uses 2--4 agents in HSSD household scenes, and covers cooperative and mixed-motive settings rather than fully adversarial ones. Future work should scale to larger teams, other domains, and deeper recursive beliefs, while further separating embodied skill from epistemic coordination for weaker models.

\para{Interpreting progress.}
A high score on \shortname{} should mean that an agent can maintain and update who-knows-what, route facts to the right partner, and act before the information becomes stale. We report literal and functional scores because belief statements alone are insufficient: literal probes test whether a belief can be stated, while functional success tests whether it is used under action and communication constraints.

\para{Using the evolving pool.}
As models improve, old tasks remain useful for historical comparability and regression checks. New rounds should be added against current frontier failures, preserving a ladder from solved to unsolved ToM behaviors. This makes \shortname{} a measurement instrument rather than a fixed leaderboard: it tracks which epistemic operations have become reliable and which still break under embodiment.

\para{Conclusion.}
\label{sec:conclusion}
We introduce \shortname{}, an evolving benchmark for functional Theory of Mind in embodied multi-agent settings. Frontier models can often \emph{state} what partners know but cannot reliably \emph{use} that knowledge during coordination. The dominant failure is epistemic coordination breakdown: withheld information, ignored partner constraints, and misallocated messages.

\bibliographystyle{unsrtnat}
\bibliography{ref}

\clearpage
\appendix
\section{Use of LLMs in this work}
\label{app:llm_use}

We used large language models as assistive tools during the development of this work, for both coding and writing. We document this usage for transparency.

\para{Coding assistance.} We used Claude Code (Anthropic) as a coding assistant throughout the development of the \shortname{} pipeline. This included writing and debugging Python code for the task generation agent (mini SWE agent), and the visualization server (which we will make public upon acceptance). The assistant helped with code refactoring and shell scripting. All generated code was reviewed and tested by the authors before integration.

\para{Writing assistance.} We used Claude Code as a writing assistant for this paper. The first drafts were manually written by the authors. The assistant helped with revising sections, shortening sentences, resolving inline comments, formatting LaTeX tables and figures, and structuring appendices. All generated text was reviewed, edited, and approved by the authors. The scientific content, experimental design, analysis, and conclusions are the authors' own.

\begin{figure}
    \centering
    \includegraphics[width=0.75\linewidth]{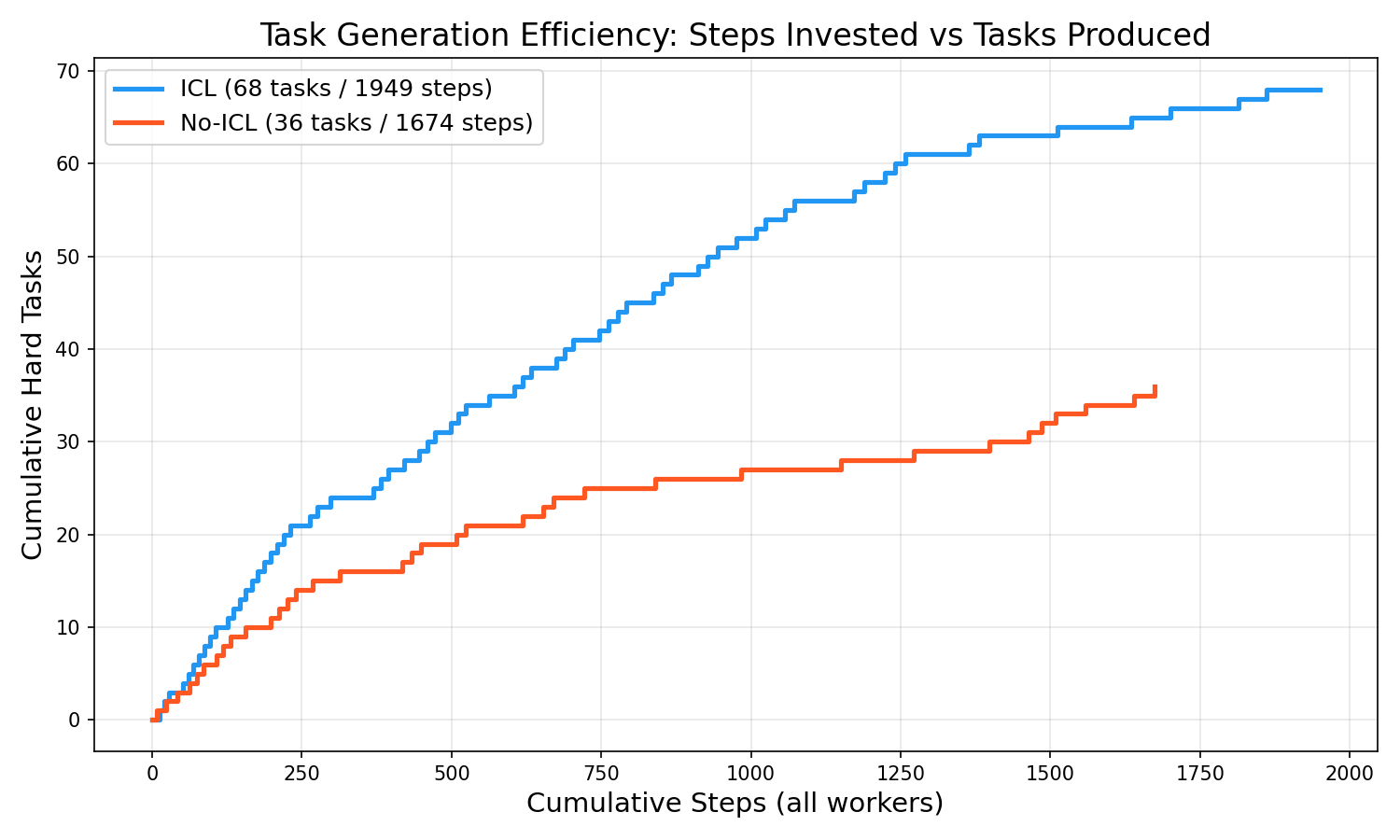}
    \caption{Cumulative tasks generated with and without ICL seed examples. With ICL (seed tasks biased toward failures), the generation agent produces accepted tasks faster and at higher epistemic depth.}
    \label{fig:icl_ablation}
\end{figure}

\section{Extended related work}
\label{app:related_work}

\para{ToM in LLMs.} Early claims that LLMs possess ToM~\citep{kosinski2024evaluating} have been challenged: trivial surface changes break performance~\citep{ullman2023large}, and apparent competence reduces to shallow heuristics~\citep{shapira2024clever, sap2022neural}. This has motivated text-based benchmarks, including ToMi~\citep{le2019revisiting}, HiToM~\citep{wu2023hi}, BigToM~\citep{gandhi2024understanding}, FANToM~\citep{kim2023fantom}, ExploreToM~\citep{exploretom2024}, OpenToM~\citep{xu2024opentom}, and multimodal variants~\citep{jin-etal-2024-mmtom, mumatom2024}. All evaluate \emph{literal} ToM~\citep{riemer2025positiontheorymindbenchmarks}: the model reads a scenario and reports beliefs. Recent work recognizes this gap: T4D~\citep{zhou2024far} shows LLMs track beliefs but fail to act on them, and SimpleToM~\citep{gu2024simpletom} confirms a stark gap between explicit inference and implicit application. These evaluations remain text-based and single-agent.

\para{Embodied multi-agent benchmarks.} Habitat~\citep{szot2021habitat}, AI2-THOR~\citep{kolve2017ai2thor}, and VirtualHome~\citep{puig2018virtualhome} provide 3D household simulators. PARTNR~\citep{chang2025partnr}, TEACh~\citep{padmakumar2022teach}, and ALFRED~\citep{shridhar2020alfred} evaluate instruction following within them. Hanabi~\citep{bard2020hanabi} and Overcooked-AI~\citep{carroll} test coordination under hidden information. SOTOPIA~\citep{zhou2024sotopia} and Generative Agents~\citep{park2023generative} evaluate social intelligence through dialogue. None formally require epistemic state reasoning: PARTNR tests instruction following, Hanabi's asymmetry is over card values not spatial beliefs, and SOTOPIA lacks embodiment.

\para{Benchmark saturation.} Static benchmarks saturate as models improve~\citep{akhtar2026saturation, ott2022mapping}, and contamination compounds the problem~\citep{jacovi2023stop, golchin2024time}. Dynamic approaches like Dynabench~\citep{kiela2021dynabench} and LiveBench~\citep{white2024livebench} use human-in-the-loop or monthly refreshes, but neither provides formal solvability or epistemic depth guarantees.

\section{ToM failure trajectory evidence}
\label{app:tom_failures}

This appendix provides trajectory evidence for each ToM failure mode identified in Section~\ref{sec:failure_behaviors}. Examples are from the manual failure audit sample and were selected from failed tasks exhibiting each pattern. Quotes are extracted directly from agent action and message fields in the benchmark logs.

\subsection{Failure to relay critical information}
\label{app:fail_relay}

\para{Example 1.} \emph{Inspection Staging with a Hidden Target Cabinet.} Cooperative, 3~agents, $\K$-depth~3. Mechanics: room restriction, restricted communication, limited bandwidth (1~message per agent).

\textbf{Agent secrets:}
\begin{itemize}[leftmargin=*]
\item agent\_0: barred from kitchen\_1. Knows the correct target for picture\_frame\_5 is cabinet\_43 in kitchen\_1. Knows fridge\_36 starts closed.
\item agent\_1: barred from closet\_1. Knows picture\_frame\_5 is on stool\_29 in closet\_1.
\item agent\_2: no private knowledge about the correct cabinet or object locations.
\end{itemize}

\textbf{What should happen:} agent\_0 should immediately tell a partner that the correct cabinet is cabinet\_43. agent\_1 or agent\_2 then retrieves the picture frame from closet\_1 and places it on cabinet\_43 in kitchen\_1.

\textbf{What actually happens:}
\begin{quote}\small
\textbf{T1--4} agent\_0 searches for objects (\texttt{FindObjectTool[picture\_frame\_5]}, \texttt{FindObjectTool[phone\_stand]}) and navigates between rooms. Does not send any message. \\
\textbf{T5} agent\_1 $\to$ agent\_2: ``Please get picture\_frame\_5 from stool\_29 in closet\_1 and place it on \textbf{cabinet\_40} in kitchen.'' Agent\_1 guesses cabinet\_40 because it does not know the correct target. \\
\textbf{T6--9} agent\_2 follows agent\_1's instruction: navigates to closet\_1, picks up picture\_frame\_5, navigates to cabinet\_40. agent\_0 continues working on other subtasks (opening chest\_of\_drawers\_34). \\
\textbf{T10} agent\_2: \texttt{Place[picture\_frame\_5, on, cabinet\_40]} $\to$ Frame placed on \textbf{wrong cabinet}. \\
\textbf{T12} agent\_2 $\to$ agent\_1: ``Picture frame placed in cabinet\_40. Confirm other tasks complete?'' \\
\textbf{T13} agent\_1 $\to$ agent\_2: ``All goals complete: picture frame on cabinet\_40, fridge\_36 open.'' Agent\_1 exhausts its message confirming wrong placement. \\
\textbf{T16} agent\_0 $\to$ agent\_2: ``Can you pick it up and place it on cabinet\_43 in kitchen\_1?'' \textbf{Too late}: agent\_2 has exhausted its messages and the episode is ending.
\end{quote}

agent\_0 held the only piece of information that determined task success (cabinet\_43) but spent 15~turns on its own physical subtasks before sending it. Agent\_0 did not model that agent\_1 would guess the cabinet and direct agent\_2 to act on that guess.

\para{Example 2.} \emph{Cabinet Confirmation Chain.} agent\_0 knows the correct cabinet is cabinet\_32. agent\_0 has one message. Instead of sharing this critical info, agent\_0 uses the message to ask about the glass location. agent\_1, lacking the cabinet info, opens cabinet\_31 (wrong) and reports it as correct.

\subsection{Epistemic chain breakdown}
\label{app:fail_kchain}

\para{Example 1.} \emph{Staging + Fridge Verification with Hidden Incentives.} 3~agents. Goal includes $\K_{a_0}(\texttt{is\_open}(\texttt{fridge\_34}))$. agent\_1 opens fridge\_34 at turn~3.

\begin{quote}\small
\textbf{T3} agent\_1: \texttt{Open[fridge\_34]} $\to$ Successful. \\
\textbf{T19} agent\_1 $\to$ all: ``Found bowl\_0 on counter, plate\_1 in cabinet\_26 \ldots Need to find bottle.'' $\to$ Mentions items but \textbf{never mentions fridge state}. \\
\textbf{T21} all agents: \texttt{Done[None]} $\to$ No agent confirmed fridge is open. K-chain not established.
\end{quote}

agent\_1 opened the fridge but never told anyone. The physical goal (fridge open) is met; the epistemic goal (agent\_0 knows it is open) is not. agent\_1 does not distinguish ``the fridge is open'' from ``my partners know the fridge is open.''

\para{Example 2.} \emph{Relay Inspection with Restricted Access.} agent\_1 holds spectacles\_1 and tells agent\_0 to place them on stand\_50 in living\_room\_1. But agent\_1 is barred from living\_room\_1 and is the one holding the item. agent\_1 never puts the spectacles down for handoff; instead places them on the wrong surface in the dining room. The physical chain breaks before the epistemic chain is even attempted.

\subsection{Private objective sabotage and disclosure}
\label{app:fail_mixed}

\para{Example 1: Sabotage.} \emph{Safety Staging with Conflicting Incentives.} Shared goal: fridge\_14 open. agent\_1's private goal: fridge\_14 closed.

\begin{quote}\small
\textbf{T14} agent\_1 $\to$ all: ``The toy fire truck is on table\_9. Can you move it to couch\_10? \textbf{I'll close the fridge after}.'' \\
\textbf{T15} agent\_1: \texttt{Close[fridge\_14]} $\to$ Directly sabotages shared goal.
\end{quote}

agent\_1 announces the sabotage before executing it. The agent does not model that teammates will read this message and could preemptively re-open the fridge.

\para{Example 2: Disclosure.} \emph{Inspection Prep with Limited Messaging.} agent\_1's private goal: fridge closed (shared goal: open).

\begin{quote}\small
\textbf{T1} agent\_1 $\to$ all: ``Cup\_4 is on cabinet\_48 in bathroom\_1. Also: \textbf{my personal objective requires the fridge CLOSED at end, conflicting with shared objective}.'' \\
\textbf{T2} agent\_2 $\to$ all: ``I can get cup\_4. But conflict: \textbf{my personal objective needs fridge CLOSED too}.''
\end{quote}

Both agents reveal conflicting private objectives in their first messages, eliminating any strategic advantage the private information could provide. Both patterns reflect a failure to reason about how one's actions change what others believe and how they will respond. The sabotaging agent does not model that teammates will detect defection. The disclosing agent does not model that revealing private information lets others preemptively block the private goal (a second- or third-order ToM failure).

\subsection{Misallocating scarce messages}
\label{app:fail_topology}

Effective communication under bandwidth constraints requires modeling who can receive a message and what each recipient needs most. When agents fail at this, they waste their limited budget on the wrong recipient or on low-priority content.

\para{Example.} \emph{One-Shot Relay (Third-Order).} 4~agents. agent\_3 knows the correct table is table\_18. agent\_3 can only message agent\_1.

\begin{quote}\small
\textbf{T8} agent\_3 $\to$ agent\_0: ``glass\_6 needs to go to table\_18'' $\to$ \textbf{Blocked}: ``You can only send messages to: agent\_1.'' \\
\textbf{T10} agent\_3 $\to$ agent\_1: ``bedroom\_2 chest needs to be closed'' $\to$ Successful, but uses message on \textbf{less critical} info. \\
\textbf{T12} agent\_3 $\to$ agent\_0: ``verify chest is closed'' $\to$ \textbf{Blocked again}.
\end{quote}

agent\_3 tries to reach agent\_0 twice (blocked both times) but never sends the table\_18 info to agent\_1 (the only reachable partner). After the first block, agent\_3 does not update its model of the communication graph. The agent neither models who it can reach nor prioritizes what its reachable partner needs most.

\subsection{Not modeling partner constraints}
\label{app:fail_constraints}

\para{Example.} \emph{Inspection Prep with Nested Confirmation.} agent\_0 is barred from kitchen\_1.

\begin{quote}\small
\textbf{T8} agent\_1 $\to$ agent\_0: ``Place bowl\_0 on counter\_29 so I can verify it's there.'' \\
\textbf{T9} agent\_1 $\to$ agent\_0: ``Agent\_0, please place bowl\_0 down on counter\_29 now.'' \\
\textbf{T10} agent\_0 $\to$ agent\_1: ``\textbf{I'm barred from kitchen.} You must place bowl\_0 on counter\_29.''
\end{quote}

agent\_1 asks agent\_0 to act in the kitchen twice before agent\_0 corrects them. agent\_1 does not model agent\_0's room restriction when delegating, wasting two scarce messages on impossible requests. In other instances of this failure mode, agents ask partners to pick up objects that the partner is already holding, or delegate tasks to agents who lack the required room access. In all cases, the constraint information is available in the shared context but the delegating agent does not consult it.

\section{Task Examples}
\label{app:examples}

\subsection{Cooperative: depth-4 epistemic chain}
\label{app:example_coop}

Four agents coordinate a ``silent inspection'' across a household: open the kitchen cabinet, close the bedroom wardrobe, and place items in designated locations. Communication follows a linear chain ($a_0 \leftrightarrow a_1 \leftrightarrow a_2 \leftrightarrow a_3$) with one message per agent. Room restrictions prevent any single agent from accessing all locations; only $a_1$ can enter the kitchen.

The PDDL goal includes:
\begin{equation}
  \K_{a_0}\bigl(\K_{a_1}\bigl(\K_{a_2}\bigl(\K_{a_3}(\texttt{is\_on\_top}(\texttt{box}, \texttt{cabinet}))\bigr)\bigr)\bigr)
  \label{eq:example_k4}
\end{equation}
This requires $a_0$ to know that $a_1$ knows that $a_2$ knows that $a_3$ knows the box is on the cabinet. The box is in the kitchen, visible only to $a_1$. For this chain to hold, $a_1$ must relay the observation through $a_2$ to $a_3$, consuming two of the chain's four message slots. Agent $a_0$ must infer that $a_1$ relayed rightward (the only viable direction), that $a_2$ forwarded it, and that $a_1$ anticipated this relay. This is third-order ToM.

\begin{table}[t]
\caption{Agent configuration for the cooperative silent inspection task. Each agent has 1 message. The linear chain and room restrictions force depth-4 epistemic reasoning.}
\label{tab:example_agents}
\begin{center}
\footnotesize
\begin{tabularx}{\linewidth}{@{} l l >{\raggedright\arraybackslash}p{0.25\linewidth} X @{}}
\toprule
\textbf{Agent} & \textbf{Can message} & \textbf{Restricted from} & \textbf{Key private knowledge} \\
\midrule
$a_0$ & $a_1$ & kitchen & Cabinet holds a cardboard box \\
$a_1$ & $a_0$ or $a_2$ & entryway, bedroom, living room & Only agent with kitchen access \\
$a_2$ & $a_1$ or $a_3$ & entryway, bathroom, kitchen & Must relay between $a_1$ and $a_3$ \\
$a_3$ & $a_2$ & kitchen & Can access bathroom (target surface) \\
\bottomrule
\end{tabularx}
\end{center}
\end{table}

% Competitive example removed from current version

\section{Epistemic compilation example}
\label{app:compilation}

We walk through the full compilation of an epistemic goal. The task has two agents (\texttt{agent\_0} and \texttt{agent\_1}), where \texttt{agent\_1} is in the same room as \texttt{bowl\_1} and \texttt{table\_22}, and \texttt{agent\_1} can communicate with \texttt{agent\_0} with a message budget of 2.

Original goal (with epistemic operators).

 \begin{lstlisting}
(and
  (is_on_top bowl_1 table_22)
  (K agent_0 (K agent_1
    (is_on_top bowl_1 table_22)))
  (is_open cabinet_34)
)
\end{lstlisting}

The planner needs to verify that (1) the bowl can be placed on the table, (2) the cabinet can be opened, and (3) \texttt{agent\_0} can come to know that \texttt{agent\_1} knows the bowl has been placed. The physical subgoals (1) and (2) are already classical \pddl{}. The epistemic subgoal (3) is not. The compiler transforms it as follows.

\para{Step 1: Create knowledge predicates.} For the leaf fact \texttt{(is\_on\_top bowl\_1 table\_22)}, the compiler creates a first-layer knowledge predicate for each agent:

\begin{lstlisting}
(knows_agent_0_a1b2c3d4) ; a0 knows fact
(knows_agent_1_a1b2c3d4) ; a1 knows fact
\end{lstlisting}

Because the goal nests two $\K$-operators, the compiler also creates a second-layer predicate for the outer agent:

\begin{lstlisting}
(knows_agent_0_e5f6g7h8) ; a0 knows a1 knows
\end{lstlisting}

\para{Step 2: Create observe operators.} \texttt{Agent\_1} is co-located with the bowl and table, so the compiler generates an observe operator that sets \texttt{agent\_1}'s knowledge predicate when the physical fact holds:

\begin{lstlisting}
(:action observe_knows_agent_1_a1b2c3d4
  :parameters ()
  :precondition (is_on_top bowl_1 table_22)
  :effect (knows_agent_1_a1b2c3d4))
\end{lstlisting}

This operator says: if the bowl is on the table and \texttt{agent\_1} is in the room (encoded in the precondition via the observability model), then \texttt{agent\_1} can be marked as knowing this fact. No real observation happens; the planner is simply checking whether a valid sequence of such operators exists.

\para{Step 3: Create inform operators.} \texttt{Agent\_1} can communicate with \texttt{agent\_0}, so the compiler generates inform operators that propagate first-layer knowledge. Each operator consumes one message token:

\begin{lstlisting}
(:action inform_a0_fact_from_a1_tok1
  :parameters ()
  :precondition (and
    (knows_agent_1_a1b2c3d4)
    (can_communicate agent_1 agent_0)
    (msg_tok_agent_1_1))
  :effect (and
    (knows_agent_0_a1b2c3d4)
    (not (msg_tok_agent_1_1))))
\end{lstlisting}

A second copy uses \texttt{msg\_tok\_agent\_1\_2}, giving the planner two opportunities (matching the budget of 2). The \texttt{(not (msg\_tok\_...))} effect consumes the token, preventing reuse.

\para{Step 4: Create nested-knowledge inform operators.} For the outer $\K$-goal, the compiler generates an operator that lets \texttt{agent\_1} inform \texttt{agent\_0} about its own knowledge state:

\begin{lstlisting}
(:action inform_a0_nested_from_a1_tok2
  :parameters ()
  :precondition (and
    (knows_agent_1_a1b2c3d4)
    (can_communicate agent_1 agent_0)
    (msg_tok_agent_1_2))
  :effect (and
    (knows_agent_0_e5f6g7h8)
    (not (msg_tok_agent_1_2))))
\end{lstlisting}

The precondition requires that \texttt{agent\_1} already knows the fact (first layer). The effect sets the second-layer predicate, establishing that \texttt{agent\_0} now knows \texttt{agent\_1} knows.

\para{Step 5: Replace the goal.} The original epistemic goal is replaced with a conjunction of classical predicates:

\begin{lstlisting}
(and
  (is_on_top bowl_1 table_22)
  (knows_agent_1_a1b2c3d4)
  (knows_agent_0_e5f6g7h8)
  (is_open cabinet_34)
)
\end{lstlisting}

\para{Step 6: Add budget tokens to the initial state.} The problem's \texttt{:init} section is augmented with tokens representing \texttt{agent\_1}'s message budget:

\begin{lstlisting}
(msg_tok_agent_1_1)
(msg_tok_agent_1_2)
\end{lstlisting}

\para{Result.} The compiled problem is entirely classical \pddl{}. Fast Downward searches over the physical actions (place, open, navigate) together with the observe and inform operators. A valid plan might be:

\begin{enumerate}[leftmargin=*, itemsep=0.15em]
\raggedright
  \item \texttt{agent\_1} places \texttt{bowl\_1} on \texttt{table\_22}.
  \item \texttt{observe\codeus knows\codeus agent\codeus 1\codeus a1b2c3d4} fires (agent\_1 sees the bowl is placed).
  \item \texttt{inform\codeus knows\codeus agent\codeus 0\codeus a1b2c3d4\codeus from\codeus agent\codeus 1\codeus tok1} fires (agent\_1 tells agent\_0 the fact, consuming token 1).
  \item \texttt{inform\codeus knows\codeus agent\codeus 0\codeus e5f6g7h8\codeus from\codeus agent\codeus 1\codeus tok2} fires (agent\_1 tells agent\_0 that it knows, consuming token 2).
  \item Another agent opens \texttt{cabinet\_34}.
\end{enumerate}

If Fast Downward finds this plan (or any valid alternative), the task is provably solvable. The $\K$-depth of 2 is read directly from the nesting structure during Step~1.

\section{Task generation agent workspace and prompt}
\label{app:workspace}

The generation agent operates in an isolated workspace directory:
\begin{lstlisting}
workspace/
  working_task.json      # task being authored
  template.json          # blank task skeleton
  current_scene.json     # rooms, furniture, objects, spawns
  sampled_tasks/         # seed tasks (biased toward failures)
  submitted_tasks/       # accepted tasks
  available_mechanics.md # mechanic registry
  available_predicates.md
  available_actions.md
\end{lstlisting}

\para{Example scene graph.} The \texttt{current\_scene.json} file describes the loaded Habitat scene. Below is an abridged example for a 2-agent, 5-room scene:

\begin{lstlisting}
{
  "scene_id": "102344280",
  "episode_id": "885",
  "rooms": ["office_1", "dining_room_1",
            "laundryroom_1", "kitchen_2",
            "entryway_1"],
  "furniture": ["table_25", "cabinet_33",
       "couch_9", "counter_30", "cabinet_31",
       "table_18"],
  "objects": ["cushion_2", "bowl_4"],
  "articulated_furniture": ["cabinet_33",
                            "cabinet_31"],
  "furniture_in_rooms": {
    "office_1":      ["table_25", "cabinet_33"],
    "dining_room_1": ["couch_9"],
    "laundryroom_1": ["counter_30"],
    "kitchen_2":     ["cabinet_31"],
    "entryway_1":    ["table_18"]
  },
  "objects_on_furniture": {
    "table_25": ["cushion_2"],
    "chair_10": ["bowl_4"]
  },
  "agent_spawns": {
    "agent_0": {"position": [1.2, 0.1, 3.4],
                "room": "office_1"},
    "agent_1": {"position": [5.6, 0.1, 2.1],
                "room": "dining_room_1"}
  }
}
\end{lstlisting}

The agent uses this to select goal-relevant objects and furniture, configure room restrictions, and ground the \pddl{} goal in actual scene entities. Only articulated furniture (cabinets, fridges, drawers) can appear in \texttt{is\_open}/\texttt{is\_closed} goals.

\para{System prompt.} The agent receives the following system prompt (abridged; full prompt is ${\sim}$400 lines):

\definecolor{promptbg}{RGB}{248, 246, 240}
\definecolor{promptframe}{RGB}{200, 195, 175}

\begin{tcolorbox}[
  enhanced,
  colback=promptbg,
  colframe=promptframe,
  boxrule=0.5pt,
  arc=3pt,
  left=5pt, right=5pt, top=5pt, bottom=5pt,
  title={\small\bfseries Generation Agent System Prompt (abridged)},
  coltitle=black,
  colbacktitle=promptbg,
  toptitle=2pt, bottomtitle=2pt,
  fonttitle=\sffamily,
  attach boxed title to top left={yshift=-2mm, xshift=4mm},
  boxed title style={colback=promptbg, colframe=promptframe, boxrule=0.5pt, arc=2pt},
  breakable,
]
\scriptsize
\ttfamily

You are a puzzle designer creating multi-agent collaboration challenges.\\[4pt]

\textnormal{\sffamily\bfseries Response format.}
Each turn: \texttt{Thought: [reasoning]} then \texttt{Action: tool\_name[argument]}.\\[4pt]

\textnormal{\sffamily\bfseries Tools.}\\
\texttt{new\_scene[N]} -- load HSSD scene with N agents, reset task.\\
\texttt{bash[cmd]} -- run shell commands (jq, cat, python3, \ldots).\\
\texttt{judge[]} -- PDDL and $\K$-depth verification + LLM quality evaluation.\\
\texttt{test\_task[]} -- physically simulate the all-secrets-public baseline.\\
\texttt{submit\_task[]} -- save task (requires judge + test\_task).\\[4pt]

\textnormal{\sffamily\bfseries Workflow.}\\
1. \texttt{new\_scene[N]} $\to$ load scene.\\
2. Inspect seed tasks in \texttt{sampled\_tasks/} for inspiration.\\
3. Edit \texttt{working\_task.json}: author the \texttt{problem\_pddl :goal} FIRST, then write \texttt{task}, \texttt{agent\_secrets}, and mechanic bindings to match. Do NOT hand-author \texttt{:objects} or \texttt{:init}.\\
4. \texttt{judge[]} $\to$ fix $\to$ repeat until pass.\\
5. \texttt{test\_task[]} $\to$ reject tasks that fail with full information.\\
6. \texttt{submit\_task[]}.\\[4pt]

\textnormal{\sffamily\bfseries Core rules.}\\
-- Author the PDDL goal as the source of truth; write narrative to match it.\\
-- Secrets state WHAT (room restrictions, target IDs, mechanic hints) but NEVER HOW (no coordination strategy, no relay instructions).\\
-- Every agent must make a distinct, non-substitutable contribution.\\
-- At least one physical action must be information-dependent: an agent cannot determine what to do without information held by another agent.\\
-- Do not prescribe coordination strategy in secrets. The agent must figure out how to communicate.\\[4pt]

\textnormal{\sffamily\bfseries Secret examples.}\\[2pt]
\textnormal{\sffamily BAD} (leaks coordination strategy):\\
{\raggedright agent\_0: "Wait for agent\_3 to tell you whether stand\_34 is open, then forward that to agent\_0."\par}\vspace{2pt}
\textnormal{\sffamily GOOD} (states constraints, agents figure out coordination):\\
{\raggedright agent\_0: ["You cannot enter hallway\_2.", "You can only message agent\_1. You can send 2 messages.", "By the end, you must be confident a teammate knows stand\_34 is open."]\par}\vspace{4pt}

\textnormal{\sffamily\bfseries Functional ToM patterns} (use at least one):\\
1. \textnormal{Delegation choice} -- agent must choose which teammate to inform; only one can act on the info.\\
2. \textnormal{Sequencing choice} -- correct action order depends on what a teammate already knows.\\
3. \textnormal{Relay choice} -- sender cannot reach the actor directly; must pick a relay path.\\
4. \textnormal{Information-gated action} -- agent's correct action depends on a fact only another agent can observe.\\
5. \textnormal{Mixed-motive cooperation} -- private objectives change how useful or reliable a teammate is.\\
%6. \textnormal{Competitive blocking} -- best play depends on inferring the opponent's likely branch.\\[4pt]

\textnormal{\sffamily\bfseries Self-tests before submitting.}\\
-- Remove all Communicate actions from the golden trajectory. Can agents still achieve all physical goals independently? If yes, the task does not test functional ToM.\\
-- Can the outermost K() agent directly walk to the room and observe the fact? If yes, the K-goal is trivial.\\
\end{tcolorbox}

\section{Judge council prompt and feedback}
\label{app:judge_prompt}

The judge council (Kimi-K2.5 and GPT-5.2) receives the following prompt (abridged):

\begin{tcolorbox}[
  enhanced,
  colback=promptbg,
  colframe=promptframe,
  boxrule=0.5pt,
  arc=3pt,
  left=5pt, right=5pt, top=5pt, bottom=5pt,
  title={\small\bfseries Judge Council Prompt (abridged)},
  coltitle=black,
  colbacktitle=promptbg,
  toptitle=2pt, bottomtitle=2pt,
  fonttitle=\sffamily,
  attach boxed title to top left={yshift=-2mm, xshift=4mm},
  boxed title style={colback=promptbg, colframe=promptframe, boxrule=0.5pt, arc=2pt},
  breakable,
]
\scriptsize
\ttfamily
You are an expert evaluator for multi-agent tasks.\\[4pt]

\textnormal{\sffamily\bfseries Context provided to the judge:}\\
-- Task category and category-specific rules\\
-- Available actions, mechanics, items, and predicates\\
-- Scene objects (rooms, furniture, objects on furniture)\\
-- The compiled formal view (PDDL after mechanic compilation)\\
-- Derived runtime semantics (functional goal + literal-ToM probes)\\
-- The full task JSON (description, secrets, mechanics, PDDL goal)\\[4pt]

\textnormal{\sffamily\bfseries Key checks:}\\
-- \texttt{task} is global and must not leak secret targets\\
-- Secrets must be actionable (exact IDs) and must not prescribe strategy\\
-- Every K() goal must be backed by a mechanic preventing direct observation\\
-- The functional projection (after dropping K-goals) must remain non-trivial\\
-- K() probes should test who knows functionally relevant facts under real asymmetry\\
-- Reward tasks where the best action depends on a partner-specific model; penalize pure fact-relay tasks\\[4pt]

\textnormal{\sffamily\bfseries Scoring:} Each criterion scored 0.0--1.0.\\[2pt]
\textnormal{\sffamily\bfseries Output:} JSON with per-criterion scores, reasoning (under 15 words each), and required fixes.\\
\end{tcolorbox}

\para{Example judge feedback.} Below is an actual judge response for a task that failed on secret quality and mechanic utilization (task selected at random from the generation logs):

\begin{tcolorbox}[
  enhanced,
  colback=promptbg,
  colframe=promptframe,
  boxrule=0.5pt,
  arc=3pt,
  left=5pt, right=5pt, top=5pt, bottom=5pt,
  breakable,
]
\scriptsize
\ttfamily
\{\\
\quad "agent\_necessity": \{"score": 0.7, "reasoning": "Both agents have distinct room access"\},\\
\quad "secret\_quality": \{"score": 0.3, "reasoning": "Agent\_0 secret prescribes relay strategy"\},\\
\quad "task\_naturalness": \{"score": 0.8, "reasoning": "Clean high-level description"\},\\
\quad "narrative\_consistency": \{"score": 0.7, "reasoning": "Matches PDDL goal"\},\\
\quad "goal\_relevance": \{"score": 0.8, "reasoning": "All conjuncts needed"\},\\
\quad "mechanic\_utilization": \{"score": 0.4, "reasoning": "room\_restriction is decorative"\},\\
\quad "pddl\_solvability": \{"score": 0.9, "reasoning": "Valid and solvable at K-2"\},\\
\quad "task\_interdependence": \{"score": 0.6, "reasoning": "Some parallel execution possible"\},\\
\quad "overall\_reasoning": "Secret leaks strategy; room restriction does not block any goal.",\\
\quad "required\_fixes": [\\
\qquad "Remove strategy hints from agent\_0 secret",\\
\qquad "Restrict agent\_0 from kitchen\_1 where cabinet\_28 is"\\
\quad ]\\
\}
\end{tcolorbox}

\section{Literal ToM probe prompt}
\label{app:literal_tom_prompt}

At the end of every episode, each agent receives the following prompt to elicit literal ToM probe answers. The probe identifiers (\texttt{k\_probe\_X}) and predicate vocabulary mirror the planner predicates introduced in Section~\ref{sec:func_tom}; one per-probe specification line is appended for each $\K$-operator extracted from $\goal$.

\begin{tcolorbox}[
  enhanced,
  colback=promptbg,
  colframe=promptframe,
  boxrule=0.5pt,
  arc=3pt,
  left=5pt, right=5pt, top=5pt, bottom=5pt,
  title={\small\bfseries Literal ToM Probe Prompt (abridged)},
  coltitle=black,
  colbacktitle=promptbg,
  toptitle=2pt, bottomtitle=2pt,
  fonttitle=\sffamily,
  attach boxed title to top left={yshift=-2mm, xshift=4mm},
  boxed title style={colback=promptbg, colframe=promptframe, boxrule=0.5pt, arc=2pt},
  breakable,
]
\scriptsize
\ttfamily
The episode is over. Do not propose any more actions.\\
Using only the episode context above, provide the requested structured report.\\
Report one structured answer per probe.\\
Respond with JSON only:\\
\{"answers":[\\
\quad\{"probe\_id":"k\_probe\_X",\\
\quad\phantom{\{}"predicate":"<predicate\_name>|unknown",\\
\quad\phantom{\{}"holds":true|false|null,\\
\quad\phantom{\{}"args":["entity\_or\_target", \ldots]\},\\
\quad\ldots\\
]\}\\
Use predicate "unknown" with holds null and empty args if the agent does not know.\\[4pt]

\textnormal{\sffamily\bfseries Allowed benchmark predicates and signatures.}\\[2pt]

\textnormal{\sffamily\textit{Spatial / Relational}}\\
-- \texttt{(is\_on\_top x:object y:furniture)} -- object is on top of furniture\\
-- \texttt{(is\_inside x:object y:furniture)} -- object is inside furniture (container)\\
-- \texttt{(is\_in\_room x:object r:room)} -- object is located in room\\
-- \texttt{(is\_on\_floor x:object)} -- object is on the floor\\
-- \texttt{(is\_next\_to x:object y:object)} -- object is adjacent to another object\\[2pt]

\textnormal{\sffamily\textit{Unary State}}\\
-- \texttt{(is\_open f:furniture)}, \texttt{(is\_closed f:furniture)} -- furniture open/closed\\
-- \texttt{(is\_clean x:object)}, \texttt{(is\_dirty x:object)} -- object clean/dirty\\
-- \texttt{(is\_filled x:object)}, \texttt{(is\_empty x:object)} -- object filled with liquid / empty\\
-- \texttt{(is\_powered\_on x:object)} -- object is powered on\\
-- \texttt{(is\_locked f:furniture)} -- furniture is locked\\[2pt]

\textnormal{\sffamily\textit{Agent}}\\
-- \texttt{(is\_held\_by x:object a:agent)} -- object is held by agent\\
-- \texttt{(agent\_in\_room a:agent r:room)} -- agent is in room\\
-- \texttt{(has\_item a:agent i:item)} -- agent has item in inventory\\
-- \texttt{(has\_at\_least a:agent i:item)} -- agent has at least N of item\\
-- \texttt{(has\_most a:agent i:item)} -- agent has the most of item among all agents\\
-- \texttt{(item\_in\_container i:item f:furniture)} -- (planner) item is hidden inside furniture until opened\\[2pt]

\textnormal{\sffamily\textit{Mechanic}} (init-only, do NOT use in \texttt{pddl\_goal})\\
-- \texttt{(is\_inverse f:furniture)} -- inverted open/close\\
-- \texttt{(mirrors f1 f2)}, \texttt{(mirrors\_closed f1 f2)} -- f1 mirrors f2's state / open-close toggle\\
-- \texttt{(controls f1 f2)}, \texttt{(controls\_unlocked f1 f2)},\\
\quad \texttt{(controls\_closed f1 f2)}, \texttt{(controls\_locks f1 f2)} -- remote control of state, unlock, open, lock\\
-- \texttt{(is\_restricted a:agent r:room)} -- agent cannot enter room\\
-- \texttt{(is\_locked\_permanent f:furniture)}, \texttt{(requires\_item f:furniture i:item)},\\
\quad \texttt{(unlocks x:object f:furniture)} -- key-gated access\\
-- \texttt{(irreversible\_enabled x:object)}, \texttt{(interaction\_locked x:object)} -- one-shot interactions\\
-- \texttt{(can\_communicate from:agent to:agent)} -- directional messaging permitted\\[4pt]

For every answer, use the exact predicate name and the exact argument order required by that predicate.\\[4pt]

\textnormal{\sffamily\bfseries Per-probe specifications (one example shown).}\\
-- \texttt{k\_probe\_1}: Predict what \texttt{agent\_2} would report about ``cabinet 31 is open''. Use ordered entities \texttt{[cabinet\_31]} and the benchmark predicate vocabulary above.
\end{tcolorbox}

\section{Key design decisions}
\label{app:design_decisions}

Several design choices in the generation pipeline were informed by failure modes observed during early development. We document each decision and the problem it addresses.

\para{Why agent necessity.} Early generated tasks often included agents that contributed nothing: an agent would be assigned to a room but have no goal-relevant object there, or two agents would have identical access and capabilities, making one redundant. When benchmarked, the redundant agent would simply idle (Wait actions for the entire episode) while the other completed the task alone. The \emph{agent necessity} criterion rejects tasks where any agent can be removed without breaking the intended solution. This forces the generator to design tasks where each agent holds unique access, information, or physical capability.

\para{Why secrets must not prescribe strategy.} In early experiments, secrets contained instructions like ``ask agent\_1 to open the fridge and relay the result to agent\_2.'' Agents followed these instructions verbatim and achieved near-perfect pass rates, but the task reduced to instruction following, not epistemic reasoning. The agent never needed to model what others know or can do; the secret told it exactly what to communicate and to whom. The \emph{secret quality} criterion now rejects any secret that leaks coordination strategy. Secrets state constraints (``you cannot enter kitchen\_1''), targets (``cabinet\_28 must end open''), and mechanic hints (``the handle is reversed''), but never the plan.

\para{Why public/private grounding.} If the shared task description $\taskdesc$ contains exact object IDs and target locations, all agents receive the same complete information and there is no reason to communicate. Information asymmetry arises only when $\taskdesc$ stays high-level (``reset the house for inspection'') and the actionable specifics (which cabinet, which room, which object) are distributed across private secrets $\Sigma(a_i)$. This split is what creates the need for agents to share information selectively.

\para{Why PDDL goal before narrative.} When the generation agent wrote the natural-language description first, it frequently invented requirements not expressible in \pddl{} (e.g., ``agents should feel satisfied with the arrangement'') or omitted requirements that were in the formal goal. Writing $\goal$ first and deriving $\taskdesc$ and $\Sigma$ from it eliminated this class of inconsistencies.

\para{Why a two-model council.} A single judge model exhibited systematic biases: GPT-5.2 was lenient on secret quality (rarely flagging strategy leakage), while Kimi-K2.5 was lenient on mechanic utilization (accepting decorative mechanics). Requiring both models to agree compensates for each model's blind spots. Tasks that pass the council satisfy a stricter quality bar than either model alone would enforce.

\para{Why baseline calibration.} The baseline condition (all secrets public) serves two purposes. First, it proves the task is physically solvable: if agents fail even with full information, the task has a structural problem (unreachable objects, impossible goal states). Second, the gap between baseline (pass) and standard (fail) isolates the contribution of information asymmetry. Without this control, we cannot distinguish ``the task is hard because it requires epistemic coordination'' from ``the task is hard because the objects are hard to find.''

\para{Why not give the planner to the evaluated agent.} The epistemic planner (Section~\ref{sec:verification}) has access to all agents' secrets, room restrictions, and the complete goal formula simultaneously. It solves the task as an omniscient coordinator. If we gave this planner as a tool to the evaluated agent, the agent could query it to determine what every other agent knows, what information to communicate, and in what order. This would bypass epistemic reasoning entirely. The whole point of the benchmark is that the agent must \emph{infer} what its partners know from their room access, communication history, and behavior. Handing the agent an oracle that answers ``what does agent\_1 know?'' would reduce the task to tool calling, not Theory of Mind.

\section{Real-world grounding of mechanics}
\label{app:mechanics_grounding}

Each mechanic in \shortname{} models a constraint that commonly arises in real-world multi-agent systems.

\para{Room restriction.} A warehouse fulfillment center assigns robots to specific zones. A robot on the packing floor cannot observe inventory levels on the storage floor. To coordinate a restock, the packing robot must communicate its needs to a storage robot that can verify shelf state directly.

\para{Limited bandwidth.} A search-and-rescue team operates on a shared radio frequency with limited airtime per responder. Each transmission must carry the most critical information first. A responder who wastes a transmission asking for confirmation of already-known facts may not have airtime left to relay a newly discovered survivor location.

\para{Restricted communication.} In a hospital, a nurse reports to the attending physician, not directly to the specialist in another department. If the specialist needs information from the nurse, the physician must relay it. The communication topology determines who can inform whom and how many hops a piece of knowledge must travel.

\para{Remote control.} A smart-home system links a wall switch in the hallway to a heater in the bedroom. The person operating the switch cannot see whether the heater actually turned on. They must either walk to the bedroom to verify, or ask someone already in the bedroom to confirm.

\para{State mirroring.} Two networked thermostats in different rooms are synchronized: adjusting one changes the setting on both. An occupant in one room who lowers the temperature may not realize they also lowered it in a room where someone else prefers it warm. Coordination requires knowing who else is affected by the shared state.

\para{Inverse state.} A pressure release valve works opposite to intuition: turning it clockwise releases pressure rather than increasing it. An operator unfamiliar with this mapping will produce the wrong effect. In a team setting, the operator who knows the mapping must communicate it to others before they interact with the valve.

\section{$\K$-depth validity study}
\label{app:kdepth_validity}

We manually verify whether the stated $\K$-depth in the \pddl{} goal matches the actual epistemic reasoning required. For each of 50 randomly sampled tasks from the mixture-optimized pool, we read the goal, agent instructions, room restrictions, and communication topology. For each $\K$-goal, we check: (a)~is the outermost agent barred from the room where the fact holds? (b)~can the inner agent directly message the outer agent, or must knowledge relay through intermediaries?

A task's $\K$-depth is \textbf{valid} if the outermost agent cannot directly observe the fact and must rely on communication or inference to learn about the inner agent's knowledge state. A task is \textbf{inflated} if the outermost agent can directly observe the inner agent performing the action (e.g., both agents are in the same room with no restriction).

\begin{table}[h]
\centering
\caption{$\K$-depth validity on 50 randomly sampled tasks. Valid: the outermost agent is barred from the fact room and must rely on communication to satisfy the $\K$-goal. Inflated: the outer agent can directly observe the fact, making the $\K$-goal trivially satisfiable without epistemic reasoning.}
\label{tab:kdepth_validity}
\small
\begin{tabular}{@{} lccc @{}}
\toprule
\textbf{$\K$-level} & \textbf{Total} & \textbf{Valid} & \textbf{Inflated} \\
\midrule
$\K$-1 & 20 & 20 (100\%) & 0 \\
$\K$-2 & 13 & 12 (92\%)  & 1 (8\%) \\
$\K$-3 & 17 & 17 (100\%) & 0 \\
\midrule
All    & 50 & 49 (98\%)  & 1 (2\%) \\
\bottomrule
\end{tabular}
\end{table}

49 of 50 tasks (98\%) have valid $\K$-depth. All $\K$-1 and $\K$-3 tasks are valid. The single inflated task is a $\K$-2 task where neither agent is restricted from the room containing the target furniture, so both can observe the state directly.

Among the valid tasks, 33 (67\%) require genuine multi-hop reasoning due to restricted communication (the inner agent cannot directly message the outer agent), while 16 (33\%) are achievable via a single direct message. In both cases, the outermost agent is barred from the fact room and must model whether the inner agent has observed the fact. The stated $\K$-depth is correct in both cases; the difference is in the communication complexity, not the epistemic depth.

\section{Compute and API cost}
\label{app:compute_cost}

All experiments run via LLM APIs, so we report wall-clock time and dollar cost in lieu of GPU-hours. Figures cover the runs reported in the paper.

\para{Task generation.} Generating one task runs the full pipeline described in Section~\ref{sec:gen_agent}: the generation agent proposes a candidate, which then passes through PDDL verification, ToM scoring, baseline calibration, and judge curation. Most candidates are filtered out, so per-accepted-task cost reflects $\sim$13$\times$ more attempts than tasks kept. Table~\ref{tab:gen_cost} reports cost at three accounting granularities; the relevant one depends on the question being asked.

\begin{table}[t]
\centering
\caption{Task-generation cost. ``Final benchmark'' counts tasks that survived the full pipeline (PDDL verification, ToM scoring, calibration, judge curation).}
\label{tab:gen_cost}
\footnotesize
\begin{tabular}{@{} lr @{}}
\toprule
\textbf{Measure} & \textbf{Cost} \\
\midrule
Per worker attempt (win or lose)        & $\sim$\$0.52 \\
Per task that passes generation         & $\sim$\$0.74 \\
Per task kept in the final benchmark         & $\sim$\$9.70 \\
\midrule
Wall-clock per attempt (mean)           & $\sim$13 min (range 6--40 min) \\
Parallel workers                        & $\sim$24 \\
\bottomrule
\end{tabular}
\end{table}

The \$0.74 figure is the raw generation cost; \$9.70 reflects the full pipeline cost, since most generated tasks are filtered out by PDDL verification, ToM scoring, calibration, and curation. As a rough rule of thumb, producing one benchmark-quality task costs \$1--10 depending on how strict the acceptance criteria are, and takes 1--2 hours of single-threaded wall-clock time end-to-end.

\para{Evaluation.} Each benchmark task is run $n{=}3$ times per model. Per-task evaluation cost varies by two orders of magnitude across the models in Table~\ref{tab:main_results}, ranging from $\sim$\$0.0020 (Gemini-Flash) to $\sim$\$0.2560 (GPT-5.4). Wall-clock per task averages $\sim$16 min.

\begin{table}[h]
\centering
\caption{Per-task evaluation cost. Each task is run $n{=}3$ times per model; cost is per single run.}
\label{tab:eval_cost}
\small
\begin{tabular}{@{} lr @{}}
\toprule
\textbf{Measure} & \textbf{Value} \\
\midrule
Cheapest model (Gemini-Flash) per task    & $\sim$\$0.0020 \\
Most expensive model (GPT-5.4) per task   & $\sim$\$0.2560 \\
Wall-clock per task (mean)                & $\sim$16 min \\
\bottomrule
\end{tabular}
\end{table}
\section{Detailed ablation studies}
\label{app:ablations}

\begin{table}[t]
\centering
\caption{Effect of removing each pipeline component.}
\label{tab:ablations}
\small
\begin{tabular}{lcc}
\toprule
\textbf{Ablation} & \textbf{Failure Rate} & \textbf{Avg. Pass Rate} \\
\midrule
Full pipeline  & \texttt{---} & 26.7\% \\
w/o baseline calibration & 51\% & \texttt{---} \\
% w/o LLM council judge & 50--90\% & \texttt{---} \\
w/o ICL seed examples & 50\% & \texttt{---} \\
w/o secret quality check & \texttt{---} & 43.5\% \\
\bottomrule
\end{tabular}
\vspace{-1em}
\end{table}

\para{Without baseline calibration.} The benchmark gate normally runs the target model in two modes: \emph{standard} (partial information) and \emph{baseline} (full information), accepting only tasks where standard fails but baseline succeeds. This ensures difficulty comes from information asymmetry, not from the task being fundamentally unsolvable. Removing the baseline run, 51\% of generated tasks are unsolvable even with full information.

\para{Without LLM council judge.} The judge scores each task on 8 quality criteria using a two-model council. A task passes only if its overall score is $\geq 0.65$ and every individual criterion scores $\geq 0.5$. Without the judge, 50--90\% of tasks fail post-hoc quality checks depending on the council model strength. A stronger judge model catches more issues: mechanics that are present but not load-bearing, secrets that leak exact object IDs, and mixed tasks whose private goals duplicate rather than conflict with shared goals.

\para{Without ICL seed examples.} Without in-context seed tasks biased toward frontier model failures, the generation agent defaults to simpler coordination patterns. The acceptance rate drops to roughly 50\% and accepted tasks have lower epistemic depth. Figure~\ref{fig:icl_ablation} shows the cumulative task yield with and without ICL.

\para{Without secret quality check.} Relaxing the secret quality constraint allows secrets to include coordination instructions (e.g., ``ask agent\_1 to open the fridge and relay the result to agent\_2''). When secrets prescribe how to coordinate, the average pass rate rises from 26.7\% to 43.5\%. Agents follow instructions verbatim without modeling partner knowledge, confirming the check is essential for measuring epistemic coordination rather than instruction following.

% \section{Model-specific behavioral analysis}
% \label{app:model_specific}
% Table~\ref{tab:model_specific_current} gives the current matched-subset behavioral breakdown for the seven reported models. The appendix mirrors the main results table: all values are hard-split overall or hard-split category averages, so they use the same denominator and fixed $n{=}3$ accounting as Table~\ref{tab:main_results}.

% \begin{table}[h]
% \centering
% \small
% \renewcommand{\arraystretch}{1.12}
% \begin{tabular}{@{} lrrrr @{}}
% \toprule
% \textbf{Model} & \textbf{Hard func. Avg} & \textbf{Hard lit. Avg} & \textbf{Lit.--Func.} & \textbf{Mixed--Coop func.} \\
% \midrule
% Gemini-Pro & 6.7 & 63.3 & +56.7 & -2.0 \\
% Gemini-Flash & 4.2 & 42.5 & +38.3 & +2.5 \\
% GPT-5.4 & 3.3 & 44.2 & +40.8 & +0.7 \\
% O3 & 5.0 & 52.5 & +47.5 & -5.7 \\
% Kimi-K2.5$^\ast$ & 5.8 & 44.2 & +38.3 & +6.2 \\
% GPT-5.4-mini$^\ast$ & 3.3 & 31.7 & +28.3 & -2.7 \\
% DeepSeek-v3.2$^\ast$ & 8.3 & 36.7 & +28.3 & +1.7 \\
% \bottomrule
% \end{tabular}
% \caption{Current model-specific behavioral breakdown on the hard split. \textit{Lit.--Func.} is hard literal Avg minus hard functional Avg; \textit{Mixed--Coop func.} is hard mixed functional Avg minus hard cooperative functional Avg. Asterisks mark partial API runs.}
% \label{tab:model_specific_current}
% \end{table}

\section{Theory of Mind in Cognitive Science} \label{app:tom-cogsci}

Theory of Mind was first posed as an empirical question by \cite{premack1978does}, who asked whether chimpanzees attribute mental states to others. The subsequent decades of research have produced a rich decomposition of the construct that informs how we evaluate it in artificial agents.

The earliest ToM precursors concern visual perspective-taking. \cite{masangkay1974early} showed that children as young as two can judge what another person sees, while \cite{flavell1977development, flavell1981young, flavell1992perspectives} introduced the Level 1 / Level 2 distinction: knowing \emph{that} someone can see something versus knowing \emph{how} it appears to them. \cite{flavell1978young} demonstrated that young children can reason about hiding objects from others --- an early form of modelling informational access. These findings ground EnactToM's use of room restrictions and partial observability as the basic mechanism for creating epistemic asymmetry.

The false-belief task \cite{wimmer1983beliefs, baroncohen1985does} became the gold standard for assessing whether an agent can represent a belief diverging from reality. \cite{wellman2001meta} established that children reliably pass around age four, and \cite{wellman2004scaling} showed that ToM capacities form a Guttman scale --- each level a prerequisite for the next, not a continuous variation --- motivating EnactToM's discrete epistemic depth levels. Second-order false belief emerges later and is substantially harder even for adults \cite{perner1985john}. A major theoretical development is the two-systems account of \cite{apperly2009two}: a fast, automatic system for tracking belief-like states \cite{onishi2005fifteen} and a slower system for deliberate propositional reasoning. Critically, these dissociate in adults --- \cite{keysar2003limits} showed that people with full access to a partner's perspective still default to egocentric interpretations under load. The dissociation between implicit tracking and explicit report that \cite{apperly2009two} describe in humans is consistent with the gap between functional and literal scores observed in our evaluation \ref{tab:main_results}, where several models score substantially higher on literal ToM probes than on functional task completion.

\cite{dennett1978beliefs} argued that the critical test of ToM is attribution of false beliefs, and \cite{dennett1987intentional} formalised the ``intentional stance'' --- predicting behaviour by attributing beliefs and rational agency. \cite{baker2009action} computationally formalised this as Bayesian inverse planning, later extended to joint inference over beliefs, desires, and percepts \cite{baker2017rational}. On the coordination side, \cite{bratman1992shared} and \cite{tomasello2005understanding} characterised shared cooperative activity as requiring mutual responsiveness and shared intentionality, while \cite{vesper2010minimal} proposed a minimal joint-action architecture built on prediction, monitoring, and coordination smoothing that does not require full recursive mentalising. EnactToM's cooperative tasks are designed to require at least this minimal architecture.

Finally, the formal apparatus for nested knowledge traces to epistemic logic \cite{aumann1976agreeing, aumann1999interactive}. The connection to strategic reasoning runs through level-$k$ models: \cite{stahl1994experimental, stahl1995experimental} and \cite{camerer2004cognitive} showed that humans reason at finite, heterogeneous depths, \cite{nagel1995unraveling} demonstrated bounded iterated reasoning in guessing games, and \cite{crawford2007fatal} applied level-$k$ analysis to spatial hide-and-seek --- structurally similar to EnactToM's embodied coordination. These models predict heterogeneous and bounded depth of reasoning, which EnactToM's per-depth evaluation is designed to measure.

\end{document}